\documentclass[preprint,12pt]{elsarticle}

\usepackage{amssymb}
\usepackage{amsmath}
\usepackage[ruled,vlined]{algorithm2e}
\usepackage{tabularx}
\usepackage{booktabs}
\usepackage{float}
\usepackage{lscape}
\usepackage{adjustbox}
\usepackage{rotating}  
\usepackage{array} 
\usepackage{comment}
\usepackage{soul}

\usepackage{xurl}
\usepackage{afterpage}
\usepackage{caption}
\usepackage{acro}        
\usepackage{pdflscape}
\usepackage{graphicx}
\usepackage{tabularx}
\usepackage{array}
\usepackage{tabularray}
\usepackage{hyperref}
\usepackage{multirow}
\usepackage[table,xcdraw]{xcolor}
\usepackage[normalem]{ulem}
\useunder{\uline}{\ul}{}
\usepackage{longtable}
\usepackage{booktabs}
\usepackage{pifont}
\usepackage{adjustbox}

\journal{Nuclear Physics B}

\begin{document}

\begin{frontmatter}



\title{Federated Explainable Artificial Intelligence: 
Roles, Architectures, Evaluation, and Open Challenges}


 %% \author{} %% Author name
\author[label1]{Masoume Gholizade}
\author[label1]{Fabrizio Ruffini}
\author[label1]{Pietro Ducange}
\author[label1]{Francesco Marcelloni}

\affiliation[label1]{organization={Department of Information Engineering, University of Pisa},
             addressline={Largo Lucio Lazzarino 1},
             city={Pisa},
             postcode={56122},
             %state={Italy},
             country={Italy}}


%% Abstract
\begin{abstract}

Federated Learning (FL) has emerged as a key paradigm for privacy-preserving collaborative model training across distributed and heterogeneous data sources. By keeping raw data local, FL addresses data confidentiality concerns, yet it does not resolve the opacity of modern machine learning models. In parallel, Explainable Artificial Intelligence (XAI) has gained attention for improving transparency, trust, and accountability, particularly in high-stakes domains. Their intersection has given rise to Federated Explainable Artificial Intelligence (FedXAI) paradigm, which aims to jointly satisfy privacy and explainability requirements.
This survey provides a systematic review of FedXAI, moving beyond the view of explainability as a purely post-hoc tool. We show how explainability is increasingly embedded as an active component of the FL lifecycle, influencing aggregation, coordination, personalization, robustness, and system-level decision making. To organize the growing literature, we introduce a multi-axis taxonomy that categorizes FedXAI methods by the role of explainability, model and explainer types, explanation scope, integration level, FL settings, and data heterogeneity.
Furthermore, we examine approaches ranging from model-agnostic explanations to interpretable-by-design federated models and explainability-aware aggregation mechanisms. We review evaluation practices and highlight the lack of standardized benchmarks and metrics for assessing explanation quality, stability, privacy leakage, and computational overhead in federated environments. Finally, we identify open challenges such as explainability under non-IID data, explanation-centric security threats, communication-efficient XAI, continual FedXAI, and the integration of domain knowledge and regulatory constraints.
By consolidating existing work and identifying key gaps, this survey serves as a reference framework for designing trustworthy, transparent, and privacy-preserving federated AI systems.

\end{abstract}

%% Keywords

\end{frontmatter}

\section{Introduction}
\label{sec1}

Federated Learning (FL) has emerged as a prominent paradigm for privacy-preserving collaborative machine learning, particularly in scenarios where centralized data collection is infeasible due to regulatory, organizational, or ethical constraints. By keeping raw data local and exchanging only model updates, FL enables learning across distributed data silos while reducing direct data exposure risks~\cite{Corcuera2022Fed-XAI}. This privacy-by-design motivation and the broader framing of FL for explainable AI models are explicitly discussed in the foundational perspective on Federated Explainable Artificial Intelligence (FedXAI)~\cite{padilla2025trustworthy}.

Despite these advantages, FL alone does not address a fundamental limitation of modern machine learning: the lack of transparency and interpretability of learned models. In practice, many real-world federated deployments still rely on black-box models, making it difficult for stakeholders to validate decisions, audit model behavior, or understand failure modes under heterogeneous clients. This issue becomes especially evident in realistic settings with non-IID data, where both model behavior and explanation consistency can vary across clients, as empirically illustrated in distributed Intrusion Detection Systems (IDS) evaluation with SHAP-based analysis~\cite{Oki2024Evaluation}.

Explainable Artificial Intelligence (XAI) is commonly used to enhance transparency by providing human-understandable explanations of predictions. However, when considered in isolation, XAI often assumes centralized access to data, model internals, or repeated querying--assumptions that conflict with the decentralized and privacy-constrained nature of FL. Moreover, explanation artifacts themselves may introduce additional privacy and security risks in federated settings; for example, FL-IDS studies highlight that explanations can become a new leakage surface and motivate secure/controlled aggregation of explanation signals~\cite{Kalakoti2025}. These limitations indicate that explainability cannot be treated as a purely post-hoc add-on, but must be redesigned to operate under federated constraints.

The need to jointly address privacy and explainability becomes most evident in high-stakes domains, where model outputs can trigger irreversible actions, regulatory consequences, or safety-critical interventions. Across the FedXAI literature, this co-design motivation repeatedly appears in healthcare, finance, IDS, and energy systems. In healthcare, FL combined with SHAP-based explanations has been used to provide clinically interpretable decision support while keeping patient data local~\cite{Kumar2025Federated}. In finance, federated fraud detection pipelines integrate SHAP to support transparency and auditability under inter-bank data silos~\cite{Transparency2024Awosika}. In cybersecurity and IDS, secure aggregation of explanation signals is leveraged to provide trustworthy interpretations without exposing raw traffic data~\cite{Kalakoti2025}. In energy systems, explainability is integrated into clustered FL to handle heterogeneity and deliver interpretable forecasting drivers (e.g., radiation and meteorological factors)~\cite{Explainable2025Ali}. Across these domains, FedXAI emerges as a natural solution in settings where data cannot be centralized and decision-making processes must remain human-validated. 

Despite the growing body of work at the intersection of FL and XAI, existing surveys remain insufficient to capture the full scope of this emerging paradigm. Prior survey-style discussions either emphasize FL/XAI separately or provide high-level conceptual framing without systematically organizing design choices, evaluation dimensions, and deployment constraints--an issue acknowledged in early FedXAI survey perspectives~\cite{Corcuera2022Fed-XAI}. Furthermore, the rapid expansion of FedXAI research has resulted in fragmented methodologies, inconsistent evaluation practices, and limited agreement on standardized benchmarks and explanation metrics. In addition, evaluation remains fragmented: metrics for explanation quality, consistency under non-IID data, privacy leakage risk from explanations, and the cost of generating/sharing explanations are not yet standardized, making cross-paper comparison difficult.

Several recent surveys have investigated FL, XAI, and their intersection from different perspectives. Early FedXAI discussions mainly provided conceptual motivations for integrating explainability into privacy-preserving federated systems~\cite{Corcuera2022Fed-XAI}. Later studies explored this intersection in domain-specific contexts. For example,~\cite{Arisdakessian2023IoTIDS} discussed FL and XAI as promising future directions for IoT intrusion detection systems, while~\cite{Chaddad2023MedicalFLXAI} focused on explainable and federated AI in healthcare and clinical decision support. Trust-oriented perspectives were further explored in~\cite{Tariq2024TrustworthyFL}, where explainability was considered alongside fairness, privacy, robustness, and accountability as components of trustworthy FL. Similarly,~\cite{Govardanan2024IoMT} reviewed the integration of FL and XAI for Internet of Medical Things (IoMT) applications, emphasizing privacy preservation and interpretability in smart healthcare systems. More recently,~\cite{dubey2025integrating} reviewed FedXAI for next-generation IoT environments, focusing on scalability, transparency, communication efficiency, and trust in decentralized systems.

Although these surveys provide valuable insights, most existing studies either focus on specific application domains or discuss explainability as a secondary trust component rather than a core design dimension throughout the FL lifecycle. Furthermore, current FedXAI research still lacks unified taxonomies, standardized evaluation protocols, and consistent benchmarking strategies under heterogeneous and privacy-constrained settings. In contrast, this survey adopts a unified taxonomy-driven and evaluation-aware perspective that systematically analyzes explainability across federated architectures, training workflows, evaluation methodologies, and deployment scenarios.

The main contributions of this survey are summarized as follows:
\begin{itemize}

   \item A multi-dimensional taxonomy is introduced to systematically categorize FedXAI methods across learning paradigms, explainability types, and integration levels within the federated pipeline.
    
    \item A lifecycle-oriented perspective is presented to analyze how explainability interacts with different stages of FL, including local training, aggregation, communication, and deployment.
    
    \item Key evaluation gaps and methodological limitations in current FedXAI studies are identified and analyzed, with particular emphasis on metrics, benchmarks, and experimental protocols under non-IID and privacy-constrained settings.
    
    \item An open challenges roadmap is synthesized, outlining critical research directions toward robust, privacy-preserving, and trustworthy FedXAI systems.
\end{itemize}

\begin{figure}[htbp]
    \centering
    \includegraphics[width=\textwidth, keepaspectratio]{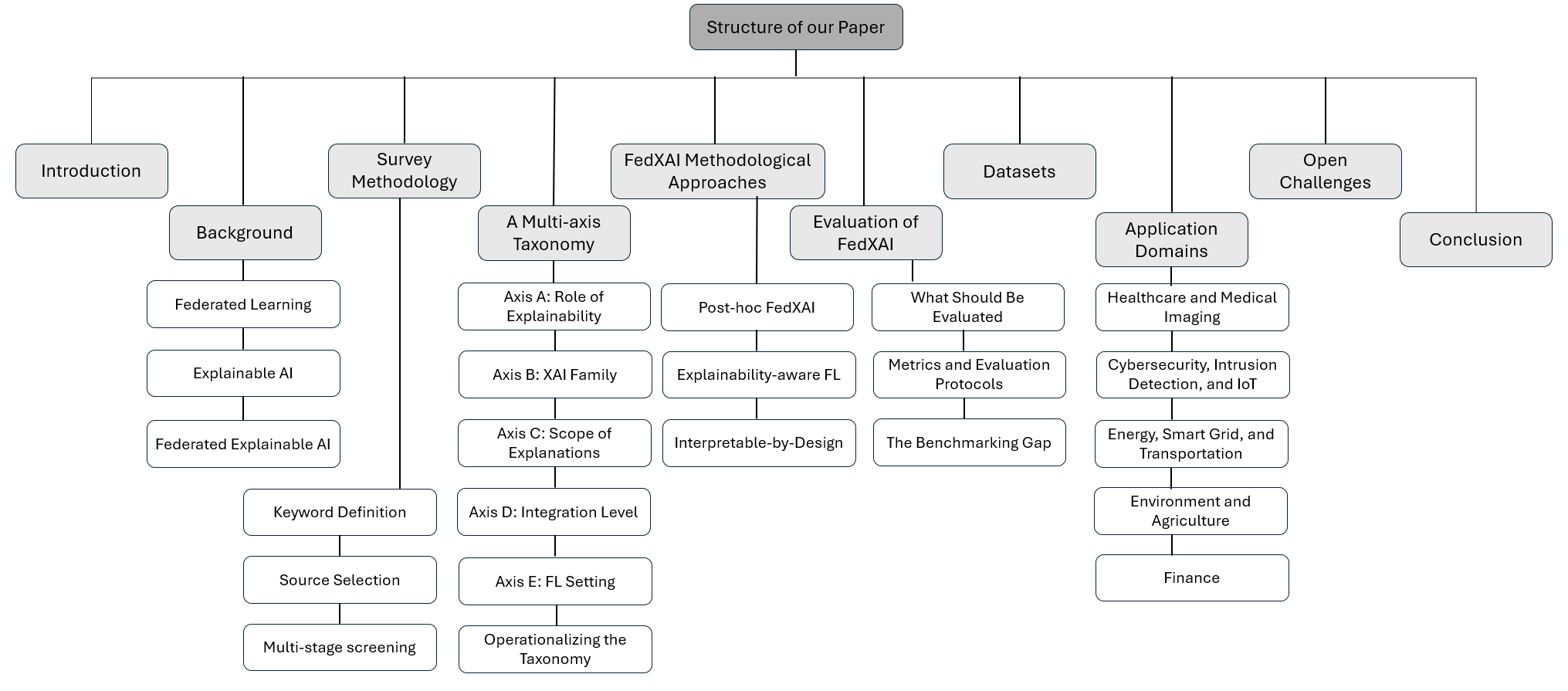}
    \caption{Conceptual and Logical Structure of the Survey.}
    \label{fig:survey_flowchart}
\end{figure}

Figure~\ref{fig:survey_flowchart} provides an overview of the logical structure of this survey. 
The proposed multi-axis taxonomy is positioned as the central element, guiding the organization of methodological approaches, evaluation criteria, datasets, and application domains. 
This structure enables a coherent and systematic analysis of FedXAI, from foundational concepts to practical deployment challenges and future research directions.

\section{ Background and Preliminaries}
\subsection{Federated Learning Fundamentals}
\label{subsec:fl_fundamentals}
FL is a distributed machine learning paradigm in which multiple clients--such as mobile devices, organizations, or edge nodes--collaboratively train a shared global model without exchanging their raw data. This concept, popularized by McMahan et al.~\cite{mcmahan2018learning}, enables learning in privacy-sensitive and geographically distributed environments by transmitting only model updates, parameters, or gradients rather than local raw data~\cite{yazdinejad2021federated}. By design, FL reduces direct risks of data exposure while still allowing models to benefit from decentralized and heterogeneous data sources.

The rapid adoption of FL is driven by both technological and regulatory factors that limit the feasibility of centralized data collection. Increasing concerns about data privacy, coupled with stringent legal frameworks such as the GDPR, restrict unrestricted data aggregation across organizations~\cite{Kumar2025Federated}. At the same time, the massive volume of data generated by personal devices, sensors, and enterprise systems makes centralized training inefficient or impractical. As a result, many real-world environments are characterized by isolated ``data silos'' that cannot be shared due to privacy, regulatory, or organizational constraints. FL addresses these challenges by enabling collaborative model training while ensuring that data remain local and under the control of their owners~\cite{Hu2024SecurityFL}.

%\paragraph{Formal FL Optimization Model.}
A standard mathematical formulation of FL is presented by Pei et al.~\cite{pei2024review}. Consider $K$ clients $\{C_1, \dots, C_K\}$, where each client $C_k$ owns a private dataset $D_k$ with $n_k$ samples. Let $n = \sum_{k=1}^K n_k$ be the total number of samples across all clients. The global learning objective is defined as:
\begin{equation}
    \min_{\omega \in \mathbb{R}^d} F(\omega) 
    = \sum_{k=1}^{K} \frac{n_k}{n} F_k(\omega),
\end{equation}
where the local empirical loss at client $k$ is given by
\begin{equation}
    F_k(\omega) = \frac{1}{n_k} \sum_{j=1}^{n_k} 
    f_{k,j}(\omega; x_{k,j}, y_{k,j}),
\end{equation}
with $f_{k,j}$ denoting the loss incurred on local sample $(x_{k,j}, y_{k,j})$. This formulation mirrors centralized empirical risk minimization, with the critical distinction that raw data remain distributed and never centrally stored or accessed.

From a performance standpoint, Blanco-Justicia et al.~\cite{blanco2021achieving} emphasize that a practical FL algorithm should achieve predictive accuracy close to that of a centralized model, i.e., a model trained under a traditional learning paradigm in which all data are centrally collected and jointly optimized:
\begin{equation}
    \left| \text{Accuracy}_{\text{fed}} - \text{Accuracy}_{\text{centralized}} \right| < \delta,
\end{equation}
where $\delta$ is a small tolerable margin. This criterion reflects the expectation that FL should preserve both privacy and competitive model performance.

%\paragraph{Federated Learning Architectures.}
Depending on the deployment context, FL systems are commonly categorized into \emph{cross-device}, \emph{cross-silo}, and \emph{hierarchical FL (HFL)} settings. Cross-device FL typically involves a large number of resource-constrained and intermittently available clients (e.g., smartphones or IoT devices), whereas cross-silo FL involves a smaller number of stable, resource-rich clients such as organizations or institutions. Hierarchical FL extends these models by introducing multi-level aggregation structures, enabling scalable coordination across edge, regional, and central servers.

%\paragraph{Statistical Heterogeneity and Non-IID Data.}
A defining challenge in FL is statistical heterogeneity across clients, commonly referred to as non–Independent and Identically Distributed (non-IID) data. This heterogeneity can arise from differences in label distributions, feature distributions, data quantity imbalance, or concept drift over time. non-IID data often leads to client drift, unstable convergence, and inconsistent global model behavior, making FL optimization substantially more challenging than centralized learning.

%\paragraph{Federated Optimization Workflow.}
The core operational workflow of FL is based on the Federated Averaging (FedAvg) algorithm~\cite{mcmahan2018learning}. Training proceeds in iterative communication rounds between a central server and a subset of participating clients. At round $t$, the server broadcasts the current global model $\omega^{(t)}$ to a subset of selected clients. Each client performs several steps of stochastic gradient descent (SGD) on its local data to compute a local update $g_k^{(t)}$, which is then sent back to the server. The server aggregates these updates using weighted averaging:
\begin{equation}
    \omega^{(t+1)} \leftarrow 
    \omega^{(t)} - \eta \sum_{k=1}^{K} 
    \frac{n_k}{n} g_k^{(t)},
\end{equation}
where $\eta$ denotes the global learning rate~\cite{McMahan2017FedAvg}. While this mechanism avoids raw data sharing, it provides only partial privacy guarantees and remains vulnerable to inference attacks on model updates~\cite{Zhu2019DLG,Melis2019Leakage}.

%\paragraph{Privacy Mechanisms in FL.}
To strengthen privacy protection, FL systems often incorporate additional mechanisms such as secure aggregation, differential privacy, homomorphic encryption, and secure multi-party computation. These techniques aim to prevent leakage of sensitive information from model updates while maintaining collaborative learning efficiency.

Overall, these principles and mechanisms have given rise to a diverse ecosystem of FL algorithms, each designed to address specific challenges such as statistical heterogeneity, communication efficiency, noisy updates, and system-level constraints. Table~\ref{tab:fl_taxonomy_landscape} provides a consolidated overview of representative FL methods, summarizing their categories, core ideas, addressed challenges, and inherent limitations.

\begin{landscape}
\begin{table*}[p]
\centering
\caption{Representative FL methods summarized along the proposed taxonomy.}
\label{tab:fl_taxonomy_landscape}
\scriptsize
\renewcommand{\arraystretch}{1.25}
\begin{tabular}{ p{2.8cm} p{2.5cm} p{7.2cm} p{3.4cm}}
\hline
 \textbf{Method} & \textbf{Category} & \textbf{Challenge--Solution Summary} & \textbf{Limitations} \\
\hline

FedAvg~\cite{mcmahan2018learning} &
Baseline FL optimization &
Addresses distributed training under data locality by iteratively aggregating local SGD updates using data-size–weighted averaging &
Highly sensitive to non-IID data; unstable convergence in heterogeneous environments \\

FedProx~\cite{Li2020FedProx} &
Optimization robustness &
Mitigates client drift caused by non-IID data by introducing a proximal regularization term that constrains local updates toward the global model &
Requires tuning an additional hyperparameter; limited benefit in near-IID settings \\

FedOptimazer~\cite{reddi2021adaptive} &
Server-side adaptive optimization &
Improves convergence stability under heterogeneity by applying adaptive optimizers (e.g., Adam, Yogi, Adagrad) at the server during aggregation &
Increases server-side computation; sensitive to optimizer parameter selection \\

FedPD~\cite{zhang2021fedpd} &
Primal--dual optimization &
Formulates FL as a saddle-point optimization problem and alternates primal and dual updates to improve convergence under heterogeneous constraints &
Higher algorithmic complexity; sensitive to dual parameter tuning \\

SlimFL~\cite{baek2022joint} &
Communication-efficient FL &
Reduces uplink and downlink communication costs by jointly optimizing split neural networks and superposition coding for compressed transmissions &
Requires compatible model partitioning; compression may degrade model accuracy \\

ConTre~\cite{chen2022contractible} &
Representation learning FL &
Alleviates feature divergence across non-IID clients by aligning latent representations via a contrastive regularization objective &
Additional computational overhead from contrastive learning; sensitive to representation dimension \\

FedGroup~\cite{duan2021fedgroup} &
Clustered FL aggregation &
Handles heterogeneous client updates by clustering clients based on similarity and performing group-wise aggregation before global updating &
Performance depends on clustering quality; cluster assignments may be unstable over time \\

FedGS~\cite{you2022reschedule} &
Gradient scheduling &
Mitigates noisy or conflicting client gradients by prioritizing and rescheduling updates based on gradient importance and stability &
Relies on heuristic ranking; may suppress infrequent yet informative updates \\

CSFedAvg~\cite{zhang2021client} &
Statistical-aware client selection &
Improves representativeness under biased participation by selecting clients according to distribution divergence metrics &
Client scoring introduces overhead; may exclude resource-constrained but informative clients \\

K-FL~\cite{kim2023k} &
Kalman-filter aggregation &
Models client updates as noisy observations and applies Kalman filtering to smooth aggregation in dynamic or noisy FL environments &
Requires accurate noise modeling; computationally expensive for high-dimensional models \\

FedOPT Framework~\cite{ahmed2024fedopt} &
Application-driven FL &
Addresses system-level heterogeneity by integrating FL with multi-layer resource scheduling in edge networks using lightweight local models &
High system complexity; domain-specific design and parameter tuning required \\

\hline
\end{tabular}
\end{table*}
\end{landscape}

\subsection{Explainable Artificial Intelligence}
\label{subsec:xai_fundamentals}

XAI encompasses a broad class of methods and frameworks aimed at mitigating the opacity of modern machine learning models and enabling systematic interpretation of their decision-making processes. While concerns about explainability have accompanied AI systems for decades, the widespread adoption of highly non-linear, high-capacity models--particularly deep neural networks--has significantly intensified the need for formal explanatory mechanisms. A notable milestone in this direction was the launch of DARPA’s XAI program, which explicitly emphasized the development of AI systems capable of articulating the reasoning behind their outputs~\cite{gunning2019darpa}. More recent surveys further highlight that increasing model complexity and deployment in high-stakes domains have elevated transparency, interpretability, and accountability from desirable properties to practical requirements~\cite{holzinger2022methods}.

Beyond technical motivations, XAI is strongly shaped by ethical and regulatory considerations. Regulatory frameworks, such as the European Union’s Artificial Intelligence Act (AI Act)~\cite{EU2024AIACT}, introduce explicit requirements for algorithmic accountability, transparency, and meaningful explanations of automated decisions, reinforcing the necessity of XAI systems~\cite{goodman2017european,mohamed2025xai}. Consequently, the overarching goal of XAI is not merely to visualize model behavior, but to transform complex decision mechanisms into representations that can be understood, scrutinized, and acted upon by human stakeholders. This subsection introduces the core conceptual distinctions in XAI that serve as the foundation for the subsequent FedXAI taxonomy and evaluation framework.

%\paragraph{Post-hoc vs.\ Interpretable-by-Design XAI.}
A fundamental distinction in XAI lies between \emph{interpretable-by-design} models and \emph{post-hoc} explanation methods~\cite{holzinger2022methods}. Interpretable-by-design approaches rely on inherently transparent model structures--such as linear models, decision trees, and fuzzy rule-based systems--where the decision logic is directly accessible. These models are particularly attractive in domains requiring immediate and unambiguous interpretability, including healthcare and fraud detection, but often face limitations in expressive power and scalability.

In contrast, post-hoc XAI methods aim to explain the predictions of already-trained black-box models without modifying their internal structure. This category includes attribution-based, surrogate-based, and visualization-based techniques that operate after model training. While post-hoc methods provide flexibility and model-agnostic applicability, they may only approximate the true internal logic of the model, potentially introducing abstraction or oversimplification in explanations~\cite{li2022extracting}.

%\paragraph{Model-Specific vs.\ Model-Agnostic Explanations.}
Another important classification distinguishes model-specific from model-agnostic explanation techniques. Model-specific methods exploit internal architectural properties of particular model families to generate explanations. For instance, Grad-CAM~\cite{selvaraju2017grad} leverages gradient information in convolutional neural networks to produce spatial heatmaps highlighting influential input regions, and has been widely applied in domains such as medical imaging and autonomous perception~\cite{singh2020explainable}. While such methods often yield high-fidelity explanations, their applicability is restricted to specific model architectures~\cite{van2022explainable}.

Model-agnostic methods, in contrast, treat the underlying model as a black box and rely solely on input-output behavior. Prominent examples include LIME~\cite{ribeiro2016should}, which approximates local decision boundaries using interpretable surrogate models, and SHAP~\cite{lundberg2017unified}, which employs cooperative game theory to assign Shapley values quantifying feature contributions. Extensions such as Anchors~\cite{ribeiro2018anchors} generate high-precision IF--THEN rules that locally anchor predictions. Although model-agnostic methods offer broad applicability, they often incur higher computational cost and may exhibit sensitivity to sampling strategies in high-dimensional settings.

%\paragraph{Instance-level vs.\ Model-level Explanations.}
XAI methods can further be categorized based on the scope of explanation. \emph{Model-level} explanations aim to characterize the overall behavior of a model across the entire dataset, providing insights into learned feature relationships, biases, and decision trends. Techniques such as model-level feature importance and Partial Dependence Plots are commonly used for this purpose~\cite{sun2024shap}. These methods are particularly useful for auditing and validation but are limited in explaining individual predictions.

In contrast, \emph{instance-level} explanations focus on individual predictions and describe how specific input features influence a particular output. Methods such as saliency maps~\cite{muller2025explainable} and counterfactual explanations~\cite{guidotti2024} enable instance-level reasoning and are especially valuable in personalized decision support and human-in-the-loop systems. In practice, model-level and instance-level explanations are often complementary rather than mutually exclusive.

In the literature, instance-level and model-level explanations are also referred to as local and global explanations, respectively. 
To avoid ambiguity with FL terminology, we use the terms instance-level and model-level explanations throughout the remainder of the paper.

%\paragraph{Explanation Quality: Fidelity, Stability, and Robustness.}
Beyond categorization, recent XAI research emphasizes the importance of evaluating explanation quality. \emph{Fidelity} measures how accurately an explanation reflects the true behavior of the underlying model. \emph{Stability} assesses the sensitivity of explanations to small perturbations in input data or model parameters, while \emph{robustness} concerns the resilience of explanations to noise, distribution shifts, or adversarial manipulation. These properties are increasingly recognized as critical, particularly when explanations are used for decision-making, auditing, or regulatory compliance~\cite{holzinger2022methods}.

%\paragraph{Implications for Federated Settings.}
While the above distinctions are well-established in centralized learning, their implications become substantially more complex in federated environments. Decentralized data access, statistical heterogeneity, and privacy constraints challenge both the generation and evaluation of explanations. These challenges motivate the need for FedXAI, where explainability mechanisms must be co-designed with FL architectures rather than treated as independent post-hoc tools. In particular, the diverse landscape of explanation paradigms used in the FedXAI papers reviewed in this survey and summarized in Table~\ref{tab:xai_taxonomy_landscape} highlights that different XAI methods impose fundamentally different requirements in terms of data accessibility, model transparency, computational cost, and privacy sensitivity. Such heterogeneity directly affects how explanations can be generated, aggregated, and validated in federated settings. The following subsection formalizes this integration and delineates the scope of FedXAI.

\begin{landscape}
\begin{table*}[p]
\centering
\caption{Representative XAI methods summarized along the proposed taxonomy.}
\label{tab:xai_taxonomy_landscape}
\scriptsize
\renewcommand{\arraystretch}{1.25}
\begin{tabular}{ p{2.8cm} p{2.5cm} p{7.2cm} p{3.4cm}}
\hline
 \textbf{Method} & \textbf{Category} & \textbf{Challenge--Solution Summary} & \textbf{Limitations} \\
\hline

Fuzzy Rule-Based Systems (TSK-FRBS)~\cite{Corcuera2022Fed-XAI,Corcuera2022,Corcuera2023Enabling,Ducange2024Federated,Corcuera2025Increasing,OpenFL2023} &
Interpretable-by-design XAI &
Provides intrinsic transparency by representing the model as human-readable fuzzy if--then rules with linguistic variables and linear consequents, enabling direct model-level interpretability &
Rule explosion in high-dimensional spaces; manual or heuristic rule design; limited scalability \\

Fuzzy Regression Trees (FRT)~\cite{Corcuera2025Increasing} &
Interpretable-by-design XAI &
Combines decision tree structures with fuzzy logic to provide interpretable hierarchical decision paths and linguistic rules while preserving competitive regression accuracy &
Model induction is specialized; limited study under adversarial or highly dynamic settings \\

LIME~\cite{Tahir2025Federated,Choir2016Sahoo,Sarker2024Enhancing,Patni2024Explainable,Taheri2025Explainable,Harshitha2025Federated,Bilal2025,Chowdhury2026FLEX-IDS,Srinivasulu2025Overcoming,Mahir2026PriFL,Aljunaid2025Secure,Weighted2025Saleem} &
Post-hoc, model-agnostic XAI &
Generates instance-level explanations by fitting simple surrogate models around individual predictions, enabling model-agnostic interpretability across diverse domains &
Low stability; sensitive to sampling strategy; explanations may lack fidelity to the true model \\

SHAP~\cite{Tahir2025Federated,Mahir2025Advanced,Explainable2025Yazdinejad,Choir2016Sahoo,Sarker2024Enhancing,Oki2024Evaluation,Explainable2025Ali,Taheri2025Explainable,Harshitha2025Federated,Ducange2024Federated,Kumar2025Federated,Fatema2025Federated,Bilal2025,Tawfik2025FedMedSecure,Corcuera2022Fed-XAI,Kalakoti2025,Aljunaid2025Secure,Amato2025,Weighted2025Saleem,Transparency2024Awosika} &
Post-hoc, model-agnostic XAI &
Uses Shapley values from cooperative game theory to fairly attribute feature contributions to predictions, supporting both instance-level and model-level explanations &
High computational cost; approximation methods may reduce accuracy and stability \\

Grad-CAM (and 1D variants)~\cite{Demystifying2024Patel,Designing2022Raza,Mastoi2025Explainable,Manocha2025Federated,Gupta2025FedMed-XAI,KbFL2025mahadi,Naz2025Privacy,Mahir2026PriFL,Towards2025Pervez} &
Post-hoc, model-specific XAI &
Produces spatial or temporal heatmaps by exploiting gradient information in convolutional layers, highlighting regions that most influence model decisions &
Restricted to CNN-like architectures; explanations can be coarse and noisy \\

Saliency and Sensitivity Maps~\cite{Mastoi2025Explainable,Towards2025Pervez} &
Post-hoc, model-specific XAI &
Visualizes gradients or perturbation sensitivity of inputs to identify influential regions or features in individual predictions &
Highly sensitive to noise; limited semantic interpretability; unstable under small perturbations \\

Integrated Gradients~\cite{Corcuera2022Fed-XAI,Kalakoti2025} &
Post-hoc, model-specific XAI &
Attributes predictions to input features by integrating gradients along a baseline-to-input path, providing axiomatic justification for attributions &
Strong dependence on baseline choice; computational overhead for deep models \\

Attention-based Explanations~\cite{Sarker2024Enhancing,Tawfik2025FedMedSecure,Srinivasulu2025Overcoming} &
Model-specific / Hybrid XAI &
Uses attention weights or cross-attention maps to highlight informative input regions or features, often combined with attribution methods for richer explanations &
Attention weights are not always faithful explanations; may be misleading without validation \\

Latent Space Visualization (e.g., UMAP)~\cite{Explainable2025Carillo} &
Visualization-based XAI &
Explains model behavior by projecting learned representations into low-dimensional spaces to analyze class separability, drift, and incremental learning dynamics &
Indirect and qualitative; sensitive to projection parameters; limited causal interpretability \\

Ontology- and Knowledge-Graph-Based Explanations~\cite{Amato2025} &
Knowledge-based XAI &
Integrates semantic concepts and domain ontologies into explanations, enabling concept-level and context-aware interpretability beyond raw feature attribution &
Requires high-quality ontologies; high computational and memory overhead; limited scalability \\

\hline
\end{tabular}
\end{table*}
\end{landscape}

\subsection{Federated Explainable Artificial Intelligence}
\label{subsec:defining_fedxai}

The increasing interaction between FL and XAI has given rise to a distinct research paradigm commonly referred to as FedXAI. 
While early studies often applied explainability techniques to federated models in an ad-hoc or post-hoc manner, more recent research frames FedXAI as a principled design space in which federated optimization and explainability mechanisms are jointly considered~\cite{Corcuera2022Fed-XAI}. 
This shift reflects growing recognition that explainability is not merely an auxiliary analysis tool, but a core component of trustworthy, privacy-preserving collaborative learning.

%\paragraph{FedXAI vs.\ XAI-under-FL.}
A critical distinction must be drawn between \emph{XAI-under-FL} and \emph{FedXAI}.  
XAI-under-FL refers to the straightforward application of existing explainability techniques to models trained via FL. In this setting, explainability is treated as an external, post-hoc analysis step that does not influence the federated optimization process itself. Typical examples include generating SHAP or LIME explanations on a trained federated model at either the client or server side, without modifying aggregation rules, communication protocols, or training dynamics. 
Several applied studies follow this paradigm, using attribution methods primarily to interpret the outcomes of federated models, for instance in intrusion detection or environmental prediction tasks~\cite{Oki2024Evaluation,Mahir2025Advanced}.

In contrast, FedXAI treats explainability as a first-class design objective that is explicitly integrated into the FL pipeline. 
Following foundational perspectives in the literature, FedXAI encompasses methods where explainability is co-designed with federated optimization, aggregation, personalization, or coordination mechanisms~\cite{Corcuera2022Fed-XAI}. 
In this paradigm, explanations are not merely outputs for human interpretation, but active signals that may guide client selection, aggregation weighting, clustering, personalization, robustness analysis, or system-level decision-making. 
Consequently, FedXAI extends beyond explaining federated models to enabling explainability-aware and explanation-guided FL.

Recent approaches exemplify this trend by incorporating explanation signals directly into aggregation or clustering strategies under non-IID data, thereby improving both transparency and learning performance~\cite{Explainable2025Ali}.

Formally, FedXAI can be characterized as a class of distributed learning systems that jointly satisfy the following properties:
(i) model training adheres to FL principles, ensuring data locality and privacy preservation;
(ii) explanation generation operates under the same decentralization and privacy constraints as model training; and
(iii) explanations are systematically incorporated into the learning, aggregation, or evaluation process rather than being confined to post-hoc analysis.

This distinction is central to the taxonomy developed in this survey, as it differentiates methods that merely \emph{explain federated models} from those that \emph{use explainability to shape FL behavior}.

%\paragraph{Threat Model for FedXAI: Explanation Leakage and Poisoning.}
While FedXAI enhances transparency and trust, it also introduces new security and privacy risks that are absent or less pronounced in conventional FL. In particular, explanation artifacts themselves may constitute sensitive information and therefore expand the attack surface of federated systems.

\emph{Explanation leakage} refers to the risk that explanations--such as feature attributions, saliency maps, or rule-based summaries--may inadvertently reveal sensitive information about local training data, data distributions, or client-specific characteristics~\cite{Corcuera2022Fed-XAI,Kalakoti2025}. Even when raw data are never shared, explanation statistics can enable inference attacks that approximate feature distributions, identify dominant attributes, or expose correlations unique to individual clients. This risk has been explicitly acknowledged in FedXAI studies employing attribution-based explanations, where explanation aggregation must be carefully controlled to avoid privacy violations~\cite{Kalakoti2025}.

\emph{Explanation poisoning} constitutes a complementary threat, wherein malicious or compromised clients deliberately manipulate explanation outputs to influence global interpretations or downstream decision-making~\cite{Chowdhury2026FLEX-IDS,Taheri2025Explainable}. In explainability-aware aggregation or clustering schemes, poisoned explanations may distort aggregation weights, bias feature relevance assessments, or mislead human operators. Unlike traditional model poisoning attacks, explanation poisoning targets the interpretability layer itself, potentially undermining trust even when predictive performance appears unaffected.

Additional threat vectors arise from inconsistencies between instance-level and model-level under non-IID data, which adversaries may exploit to amplify confusion or conceal malicious behavior. These risks highlight that explanation integrity, confidentiality, and robustness must be explicitly considered in FedXAI system design.

%\paragraph{Implications for FedXAI Design.}
The above distinctions and threat considerations underline that FedXAI cannot be realized by naively porting centralized XAI techniques into federated settings. Instead, explanation generation, aggregation, and evaluation must be aligned with federated constraints, adversarial assumptions, and communication budgets. Secure aggregation of explanation signals, robustness-aware explanation metrics, and privacy-preserving explanation mechanisms therefore emerge as essential components of FedXAI architectures.

These considerations motivate the multi-axis taxonomy introduced in the next section, which systematically categorizes FedXAI methods based on how explainability is integrated into the FL lifecycle, the type and scope of explanations produced, the role of explanation signals in training or aggregation, and the federated setting under which they operate.

\section{Survey Methodology}
\label{sec:survey_methodology}

\begin{figure}[htbp]
\centering
\includegraphics[width=0.95\textwidth]{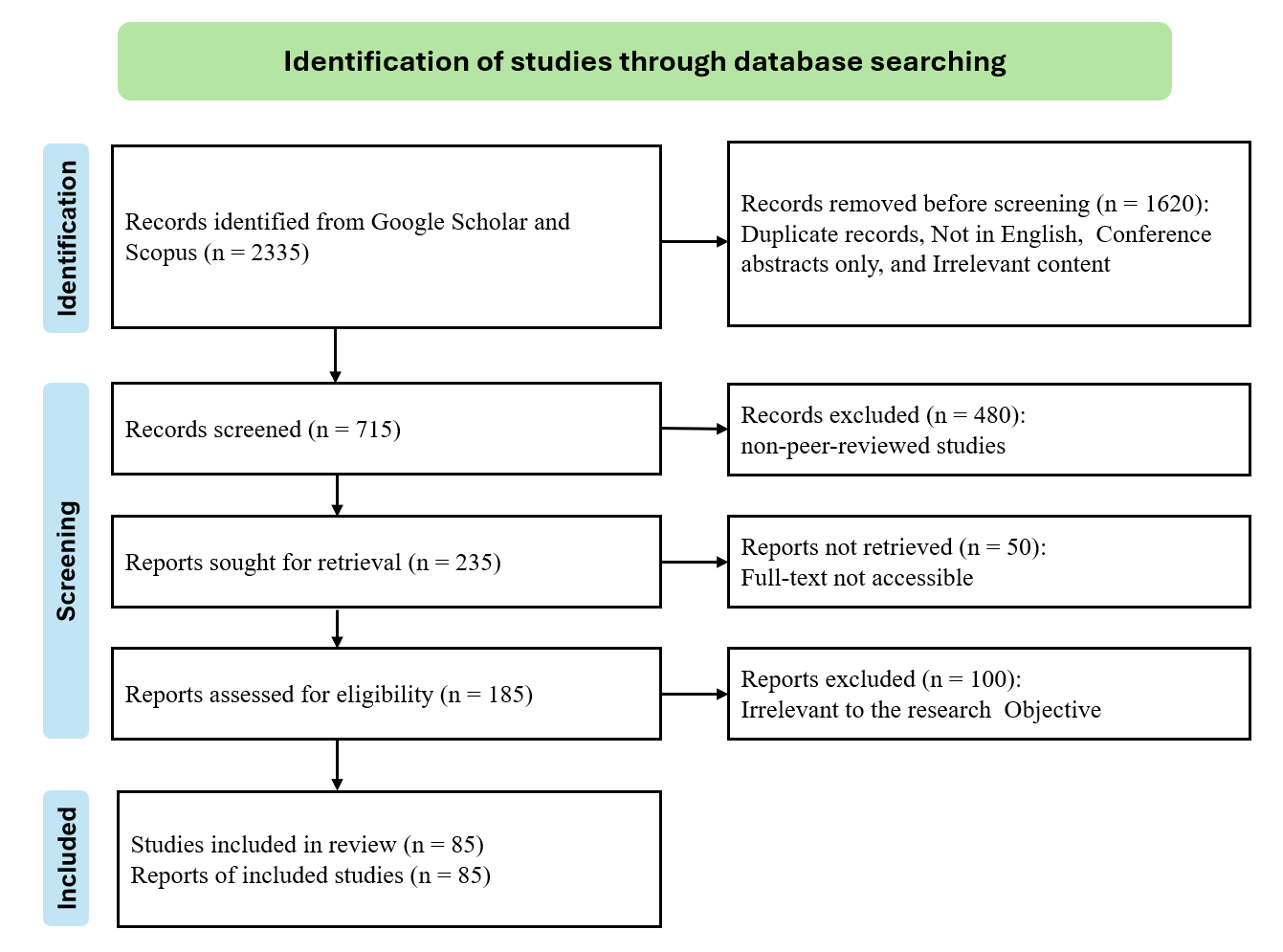}
\caption{PRISMA flow diagram of the literature selection process.}
\label{fig:prisma}
\end{figure}

This survey is conducted as a systematic literature review following a PRISMA-inspired methodology and established systematic review guidelines~\cite{Okoli2015,Kitchenham2007} to ensure transparency, reproducibility, and methodological rigor. The overall workflow consists of three main stages: (i) keyword definition, (ii) source selection, and (iii) multi-stage screening and eligibility assessment. The study selection process is illustrated in Fig.~\ref{fig:prisma}.

\subsection{Keyword Definition}

A structured search strategy was developed to comprehensively cover the intersection of FL and XAI. The queries combined ``federated learning'' with major explainability-related concepts, including ``federated explainable AI'', ``explainable federated learning'', ``interpretable federated learning'', ``federated XAI'', and ``interpretable machine learning''.

To address terminological variations, additional terms such as ``trustworthy federated learning'', ``privacy-aware explainability'', and ``federated model interpretation'' were incorporated. All keywords were combined using Boolean operators (AND, OR) and applied to titles, abstracts, and keywords across databases.

\subsection{Source Selection}

The literature search was conducted across Google Scholar and Scopus to ensure broad coverage and maximize recall. Multiple keyword combinations were applied, resulting in 2335 records prior to screening.

After removing duplicate, non-English, conference-abstract-only, and irrelevant records, 715 studies remained for the screening phase. All searches were performed up to March 2026. Most relevant papers were published after 2017, indicating that FedXAI is an emerging research area. The complete study selection process is illustrated in Fig.~\ref{fig:prisma}.

\subsection{Multi-stage screening and eligibility assessment}

A multi-stage screening process was conducted, including title/abstract filtering and full-text assessment. During the screening stage, 480 non-peer-reviewed studies were excluded, resulting in 235 reports sought for retrieval. Among them, 50 reports could not be retrieved due to inaccessible full texts.

Subsequently, 185 reports were assessed for eligibility. Studies were included if they explicitly integrated FL with explainability mechanisms, proposed explainability-aware federated frameworks, or investigated interpretable and transparent learning strategies within federated environments.

Studies focusing solely on FL or XAI, as well as tutorials, editorials, duplicate records, and non-peer-reviewed works, were excluded. After full-text assessment, 100 studies were excluded because they were not directly relevant to the research objectives. Following the screening and eligibility phases, 85 studies were selected for the final analysis. Backward snowballing was also performed to ensure completeness of the review.

A qualitative assessment was conducted to verify the methodological soundness, experimental validity, and reproducibility of the selected studies.
\section{A Multi-axis Taxonomy for FedXAI}
\label{Multi-axisTaxonomy}

To systematically organize and analyze the rapidly expanding literature at the intersection of FL and XAI, we propose a \emph{multi-axis taxonomy} for FedXAI.
Unlike conventional classifications that focus solely on explanation methods, the proposed taxonomy captures multiple complementary dimensions of explainability in federated systems, including \emph{how} explanations are generated, \emph{why} they are employed, \emph{where} they are integrated within the federated pipeline, and \emph{under which federated conditions} they operate.

Concretely, each FedXAI approach is characterized along five orthogonal axes:
(A) the role of explainability within the FL lifecycle,
(B) the family of XAI techniques employed,
(C) the scope and target of explanations,
(D) the integration level and the nature of shared artifacts, and
(E) the FL setting and heterogeneity assumptions.

Together, these axes define a structured design space for analyzing, comparing, and designing FedXAI systems across diverse explainability objectives and federated settings. The practical use of the taxonomy for system design is further discussed in Section~\ref{subsec:taxonomy_operationalization}.

\begin{figure}[htbp]
    \centering
    \includegraphics[width=0.8\textwidth]{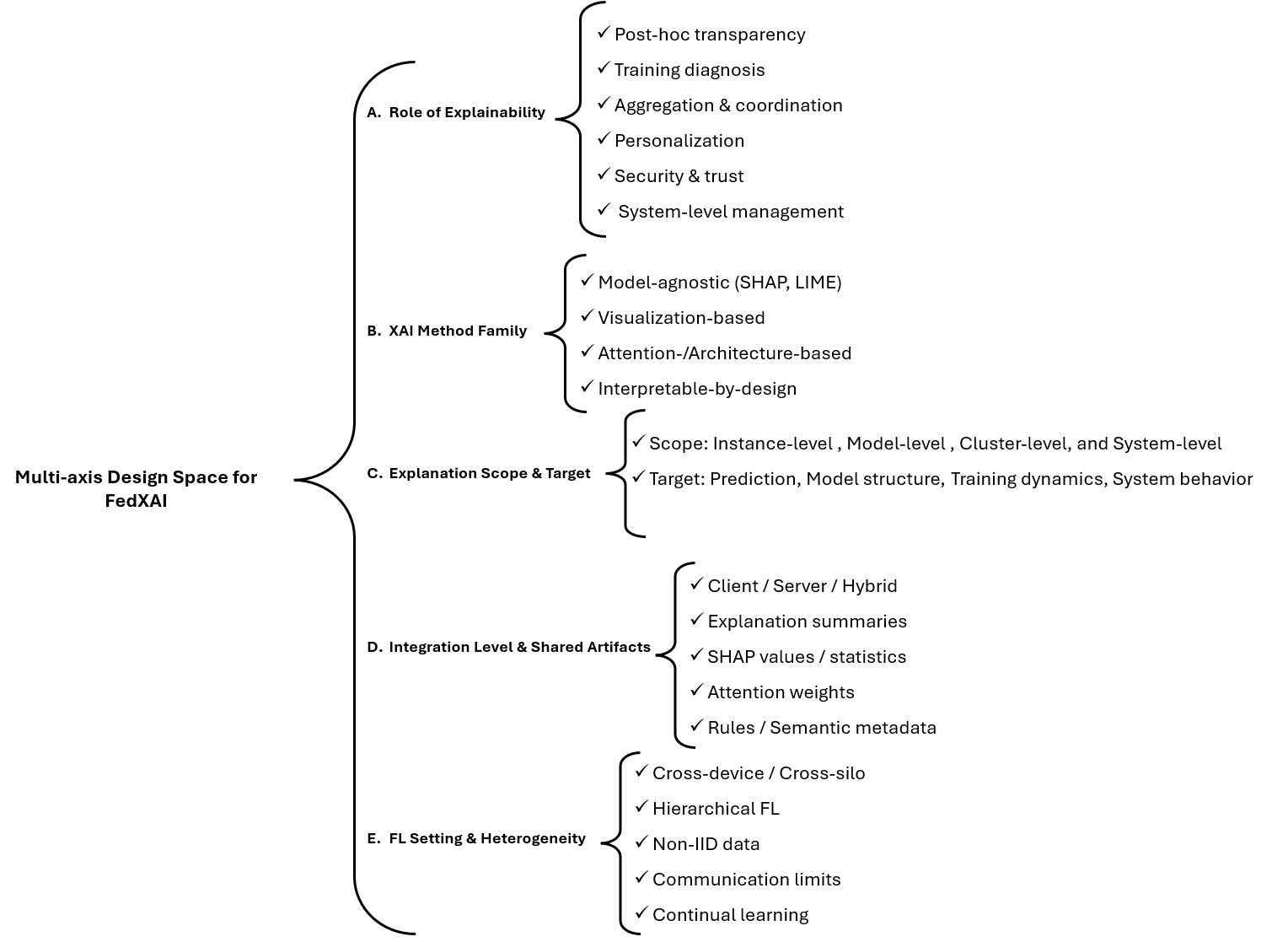}
    \caption{Proposed multi-axis taxonomy for FedXAI, illustrating the main design dimensions of explainable FL systems.}  
    \label{fig:fedxai_taxonomy}
\end{figure}

An overview of the proposed taxonomy and its categorical dimensions is summarized in Figure~\ref{fig:fedxai_taxonomy}.

\subsection{Axis A: Role of Explainability in the FL Lifecycle}

Axis~A captures \emph{why} explainability is introduced in a federated system.
Unlike centralized XAI, where explanations are typically post-hoc and user-facing, FedXAI exhibits a spectrum of roles ranging from transparency to system-level coordination and security.
From a design perspective, this axis helps practitioners identify the primary purpose of explainability in their system: post-hoc transparency, training diagnosis, aggregation and coordination, personalization, or security and trust management.

In transparency, explainability is applied after federated training to interpret model predictions without influencing optimization or aggregation.
This is the most common and least invasive form of FedXAI.

For instance, Oki \emph{et al.}~\cite{Oki2024Evaluation} apply SHAP to analyze how federated training recovers intrusion detection performance under non-IID data, comparing feature relevance between federated and centralized settings.
Similarly, Kumar \emph{et al.}~\cite{Kumar2025Federated} employ SHAP to provide clinically interpretable explanations for federated liver disease prediction while preserving institutional data privacy.
Here, explainability functions as an observability mechanism to analyze training dynamics, non-IID effects, and model evolution across rounds or episodes.

Carillo \emph{et al.}~\cite{Explainable2025Carillo} integrate attribution stability analysis and latent-space visualization into federated class-incremental learning, enabling diagnosis of catastrophic forgetting and bias correction behavior.
At the system level, Patni and Lee~\cite{Patni2024Explainable} use XAI to explain resource utilization and communication bottlenecks in hierarchical FL.
In aggregation-centric FedXAI, explanation signals actively influence coordination mechanisms such as client weighting or clustering.

Ali \emph{et al.}~\cite{Explainable2025Ali} propose SHAP-driven clustered FL for solar forecasting, where feature attributions determine cluster formation and aggregation.
In a different direction, SemFedXAI~\cite{Amato2025} introduces ontology-guided semantic aggregation, using knowledge-based explanations to align heterogeneous healthcare clients.
Personalization-oriented FedXAI leverages explainability to adapt models or explanations to client-specific contexts.

\begin{figure}[htbp]
    \centering
    \includegraphics[width=0.6\textwidth]{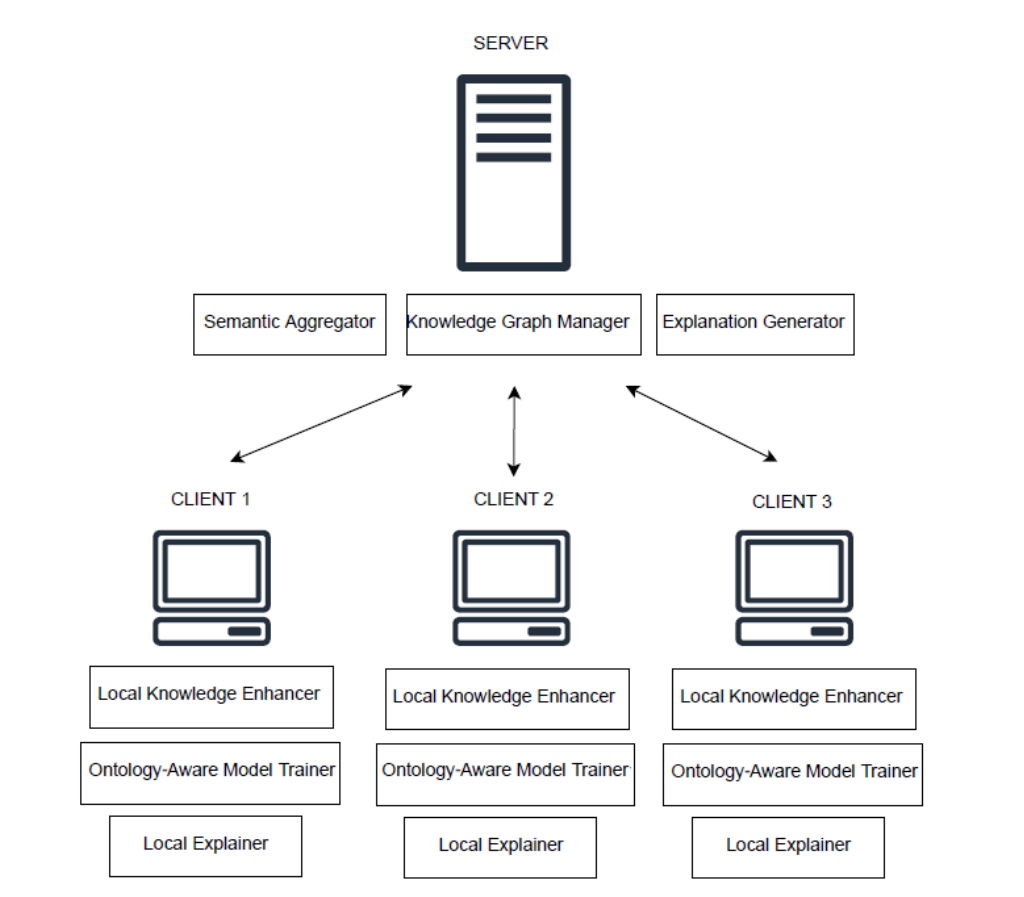}
    \caption{ General architecture of the SemFedXAI framework, which extends the traditional client
server architecture of FL with both server-side and client-side semantic components~\cite{Amato2025}.}
    \label{fig:semfedxai_architecture}
\end{figure}

Figure~\ref{fig:semfedxai_architecture} illustrates the architecture of the SemFedXAI framework~\cite{Amato2025}, which represents a canonical example of explanation-driven aggregation in FL.
Unlike conventional FedAvg-based pipelines, SemFedXAI integrates domain ontologies and knowledge graphs into both the explanation and aggregation processes.
Clients enrich feature representations using semantic concepts, generate ontology-aware explanations, and transmit semantically weighted updates to the server.
The central aggregator then performs semantic aggregation, aligning heterogeneous client updates based on shared medical concepts rather than purely numerical similarity.
This example illustrates how axis~A can be operationalized: when the design goal is coordination across heterogeneous clients, explainability is not only used for interpretation but also becomes part of the aggregation logic.

Taheri \emph{et al.}~\cite{Taheri2025Explainable} integrate SHAP and LIME into federated intrusion detection for connected vehicles, supporting adaptive local models under highly dynamic data distributions.
FedMedSecure~\cite{Tawfik2025FedMedSecure} further exploits attention weights and prototype explanations to enable few-shot personalization in federated healthcare cybersecurity.
In this role, explainability supports attack detection, trust assessment, and explanation integrity.

Yazdinejad \emph{et al.}~\cite{Explainable2025Yazdinejad} combine SHAP with differential privacy and homomorphic encryption to explain threat detection decisions while mitigating information leakage.
Kalakoti \emph{et al.}~\cite{Kalakoti2025} address explanation confidentiality by securely aggregating client-side SHAP values using secure multiparty computation.
Thus, axis~A provides an initial design guideline: practitioners should select the role of explainability according to the intended system objective, before choosing a specific XAI technique or integration strategy.

\subsection{Axis B: XAI Family}
\label{subsec:axisB_xai_family}

Axis~B categorizes FedXAI approaches based on the \emph{family of explainability techniques} employed, independently of the FL topology or application domain.
This axis captures \emph{how explanations are generated} and reflects a fundamental design choice that directly impacts explanation fidelity, computational cost, communication overhead, and compatibility with federated constraints.
From a system design perspective, this axis helps practitioners select explanation mechanisms according to model complexity, resource availability, privacy requirements, and interpretability needs.

In the surveyed literature, FedXAI methods predominantly fall into four explainability families: (i) model-agnostic post-hoc explainers, (ii) visualization-based techniques, (iii) attention- or architecture-based mechanisms, and (iv) interpretable-by-design models.
Each family exhibits distinct strengths and limitations when deployed in decentralized and privacy-preserving environments.
%\paragraph{B1) Model-agnostic post-hoc explainers.}
Model-agnostic explainability techniques constitute the most widely adopted XAI family in FedXAI.
These methods treat the trained federated model as a black box and derive explanations by probing input--output relationships, making them readily applicable to heterogeneous models and clients.

In IoT-centric FedXAI systems, explainability is typically applied in a post-hoc manner and does not interfere with federated optimization. A representative example is the FL-XAI framework for malicious traffic detection in IoT networks proposed by Bilal et al.~\cite{Bilal2025}, where a global model is trained using standard FedAvg aggregation, and interpretability is achieved through instance-level SHAP attributions computed on the trained federated model.

Such post-hoc explainers are particularly suitable for resource-constrained or heterogeneous FL settings because they can be integrated without modifying the underlying federated optimization pipeline.

SHAP and LIME dominate this category, particularly in tabular and time-series domains such as intrusion detection, fraud detection, and healthcare analytics.
For example, Oki \emph{et al.}~\cite{Oki2024Evaluation} employ Kernel SHAP to analyze how FL alters feature importance distributions in distributed intrusion detection systems, demonstrating convergence toward centralized feature relevance under FL.
Similarly, Awosika \emph{et al.}~\cite{Transparency2024Awosika} integrate SHAP-based explanations into federated financial fraud detection to provide both model-level and transaction-level transparency while preserving inter-bank data privacy.

The popularity of model-agnostic explainers stems from their flexibility and minimal assumptions about model structure.
However, in federated settings they often introduce substantial computational overhead and may raise privacy concerns when explanation statistics are shared or aggregated across clients.
%\paragraph{B2) Visualization-based explainability.}
Visualization-based XAI techniques are predominantly used in FedXAI applications involving high-dimensional spatial or temporal data, most notably medical imaging and signal analysis. These methods generate saliency maps or activation heatmaps that highlight regions of the input contributing most strongly to a prediction.

Grad-CAM and related saliency techniques are extensively employed in federated medical imaging. Mastoi \emph{et al.}~\cite{Mastoi2025Explainable} apply Grad-CAM and saliency maps to explain federated brain tumor classification from MRI images, enabling clinicians to verify that models focus on clinically relevant tumor regions despite decentralized training. 
Grad-CAM saliency maps highlight clinically relevant tumor regions in MRI images.
In hierarchical FL for leukemia diagnosis, Pervez \emph{et al.}~\cite{Towards2025Pervez} combine saliency maps, occlusion sensitivity, and RISE to evaluate explanation fidelity across different federation levels.

Visualization-based explainers offer intuitive and domain-aligned interpretations, particularly for expert users.
Nevertheless, their explanations are often qualitative, model-specific, and difficult to aggregate or compare across clients, which limits their direct use in explainability-driven coordination or optimization.
Accordingly, this family is particularly appropriate for high-stakes domains such as healthcare, where visual interpretability and clinician trust are prioritized over lightweight deployment.

% \paragraph{B3) Attention- and architecture-based explainability.}

Attention-based and architecture-integrated explainability methods derive interpretability directly from model components such as attention weights, feature-routing mechanisms, or prototype similarities.
In contrast to post-hoc methods, these approaches embed explainability into the learning architecture itself.

In smart grid load forecasting, Sarker \emph{et al.}~\cite{Sarker2024Enhancing} integrate attention mechanisms within a federated deep learning model, using attention weights to highlight influential temporal and contextual features driving energy demand predictions.
Similarly, FedMedSecure~\cite{Tawfik2025FedMedSecure} leverages cross-attention mechanisms and prototype similarity in a federated few-shot learning framework, providing interpretable insights into feature relevance and decision confidence under extreme data scarcity.

Architecture-based explainability typically offers lower runtime overhead at inference time and tighter coupling between learning and explanation.
However, interpretability is often indirect and requires careful validation to ensure that attention weights or internal activations faithfully reflect causal importance, particularly in non-IID federated environments.
These methods are therefore well suited for personalization-oriented or adaptive FedXAI systems, where explanation signals are tightly coupled with model behavior.

%\paragraph{B4) Interpretable-by-design models.}
Interpretable-by-design approaches pursue transparency through inherently explainable model structures, such as decision trees, rule-based systems, and fuzzy models.
In FedXAI, this family represents the strongest form of integration between learning and explainability, as explanations are intrinsic to the trained model rather than generated post hoc.

Corcuera Bárcena \emph{et al.}~\cite{Corcuera2022} propose an FL framework for Takagi--Sugeno--Kang fuzzy regression models, where interpretable fuzzy rules are learned locally and aggregated centrally without exchanging raw data.
Extending this line of work, federated fuzzy regression trees~\cite{Corcuera2025Increasing} enable privacy-preserving induction of transparent tree structures through the exchange of sufficient statistics, achieving performance comparable to centralized learning while retaining full interpretability.

Interpretable-by-design FedXAI methods provide high transparency, low explanation ambiguity, and reduced risk of explanation leakage.
However, they often require customized aggregation schemes and may sacrifice predictive performance or scalability when compared to deep neural models.
Consequently, they are most appropriate in applications requiring strong transparency guarantees and low explanation ambiguity, even at the cost of reduced model flexibility.

\subsection{Axis C: Scope and Target of Explanations}
\label{subsec:axisC_scope}

Axis~C characterizes FedXAI methods based on the granularity (scope) at which explanations are generated and the explanation target.
This distinction is particularly important in federated settings, where explanations may serve different stakeholders (clients, servers, system operators) and may operate at different structural levels of the learning process.
From an operational perspective, this axis helps determine whether explanations should support individual decision interpretation, global model auditing, client-group coordination, or system-level monitoring.

%\paragraph{Scope of explanations.}
The scope defines the level at which explanations are generated and interpreted, ranging from individual predictions to aggregated or cluster-level behavior.
\emph{Instance-level explanations} focus on individual predictions or instances at a specific client.
For example, Raza \emph{et al.}~\cite{Designing2022Raza} employ Grad-CAM adapted to 1D ECG signals to provide patient-specific explanations for federated arrhythmia classification.
These explanations remain confined to each healthcare institution and are intended for clinician-level interpretation.

\emph{Model-level explanations} summarize the overall behavior of the federated model.
Oki \emph{et al.}~\cite{Oki2024Evaluation} generate global SHAP summaries to compare feature importance between federated and non-distributed intrusion detection systems,
illustrating how federated aggregation reshapes model-level decision logic under non-IID data.

\emph{Cluster-level explanations} occupy an intermediate granularity, explaining the behavior of groups of similar clients.
Ali \emph{et al.}~\cite{Explainable2025Ali} use SHAP values to form client clusters in solar energy forecasting, where explanations characterize cluster-specific feature relevance and directly guide cluster-wise aggregation.

Accordingly, practitioners may select instance-level explanations for personalized decision support, model-level explanations for auditing and transparency, or cluster-level explanations for coordination and adaptive aggregation.

%\paragraph{Target of explanations.}
The target specifies which component of the federated system is being explained. Many FedXAI works target \emph{model predictions}, particularly in high-stakes applications such as fraud detection and intrusion detection.
For instance, Awosika \emph{et al.}~\cite{Transparency2024Awosika} use SHAP to explain transaction-level fraud predictions produced by a federated deep learning model, supporting regulatory transparency.

Other approaches focus on \emph{model parameters or symbolic structures}.
Corcuera Bárcena \emph{et al.}~\cite{Corcuera2025Increasing} explain federated fuzzy regression trees by exposing learned rules, splits, and linguistic terms, enabling direct inspection of the global model structure rather than post-hoc attribution.

Explainability may also target \emph{training dynamics}.
Carillo \emph{et al.}~\cite{Explainable2025Carillo} analyze attribution stability and latent representations across federated incremental learning episodes, using explanations to understand catastrophic forgetting and bias evolution over time.

Finally, some works target \emph{system-level resources and coordination decisions}.
Patni and Lee~\cite{Patni2024Explainable} apply SHAP and LIME to explain resource allocation, communication cost, and scheduling decisions in hierarchical FL, shifting the focus of explainability from prediction outcomes to system behavior. 
This paradigm is explicitly illustrated in their framework design, where an explainable AI-powered resource management unit is embedded into the core control workflow and directly influences scheduling, aggregation, and communication efficiency.

In conclusion, axis~C provides guidance on selecting both the granularity and the target of explanations according to the intended stakeholder and operational objective of the federated system.

\subsection{Axis D: Integration Level and Shared Artifacts}
\label{subsec:axisD_integration}

Axis~D captures \emph{where} explainability is integrated within the federated architecture and \emph{which artifacts are exchanged} between participants.
This axis is critical for understanding privacy implications, communication overhead, and trust boundaries in FedXAI systems.
From a practical perspective, axis~D helps determine how explanations should be distributed across clients and servers, and which explanatory artifacts can be safely exchanged under privacy and communication constraints.

%\paragraph{Client-side integration.}
In client-centric FedXAI, explanations are generated locally and are not shared with the server.
This design maximizes privacy and aligns with human-in-the-loop use cases.

For example, Raza \emph{et al.}~\cite{Designing2022Raza} generate Grad-CAM explanations locally at healthcare institutions for ECG classification, ensuring that both data and explanations remain confined to the client.
The server only aggregates model parameters, without access to explanations or raw signals.

Client-side integration is therefore particularly suitable for privacy-sensitive applications where explanations are intended primarily for local users or domain experts.
%
%\paragraph{Server-side integration.}
In server-centric approaches, explanations are generated or aggregated at the central coordinator, providing a model-level view of model behavior.

Awosika \emph{et al.}~\cite{Transparency2024Awosika} compute model-level SHAP summaries at the server to explain federated fraud detection models trained across multiple banks.
This setup supports regulatory auditing but requires careful consideration of information leakage from aggregated explanations.
Such server-side strategies are more appropriate when global auditing, monitoring, or regulatory transparency is prioritized over strict local confidentiality.

%\paragraph{Hybrid integration.}
Hybrid FedXAI combines client-side explanation generation with secure aggregation or selective sharing mechanisms, balancing privacy and model-level interpretability.
Kalakoti \emph{et al.}~\cite{Kalakoti2025} propose a hybrid architecture in which clients compute SHAP values locally and securely aggregate them using secure multiparty computation.
This enables server-side explanations that closely approximate centralized SHAP results without exposing client-level explanations or data.
Hybrid approaches provide a practical compromise between local privacy preservation and model-level interpretable system behavior.

%\paragraph{Shared artifacts.}
Across these integration levels, the nature of shared artifacts varies substantially.
Standard FL exchanges gradients or model weights, whereas FedXAI may additionally share SHAP statistics, explanation summaries, fuzzy rules, attention weights, or semantic metadata.
For instance, fuzzy rule-based FedXAI~\cite{Corcuera2025Increasing} exchanges symbolic rule parameters, while SemFedXAI~\cite{Amato2025} incorporates ontology-derived semantic information into aggregation.

In conclusion, axis~D provides guidance on selecting both the integration locality and the type of exchanged explanatory artifacts according to system-level privacy, communication, and trust requirements.

\subsection{Axis E: FL Setting and Heterogeneity}
\label{subsec:axisE_setting}

Axis~E situates FedXAI methods within specific FL settings and heterogeneity assumptions.
The feasibility, reliability, and interpretation of explanations are strongly influenced by client scale, availability, and data distribution.
This axis helps practitioners align explainability strategies with the operational characteristics and constraints of the underlying federated environment.

%\paragraph{Cross-device federated learning.}
In cross-device settings, a large number of resource-constrained and intermittently available clients participate in training.

Bilal \emph{et al.}~\cite{Bilal2025} study explainable federated intrusion detection in IoT networks, where lightweight models and SHAP/LIME explanations must operate under severe communication and computational constraints.
In such environments, explanation latency and overhead are critical design considerations.
Accordingly, cross-device FedXAI systems often favor lightweight and low-overhead explainability mechanisms.

%\paragraph{Cross-silo federated learning.}
Cross-silo FL involves a smaller number of stable, institution-level clients, making it well-suited for explainability-intensive applications.
Medical imaging~\cite{Mastoi2025Explainable} and healthcare prediction~\cite{Kumar2025Federated} exemplify this setting, where Grad-CAM and SHAP explanations are used to support clinical trust and regulatory compliance.
The relative stability of clients enables richer explanations and more complex models.

This setting is therefore more compatible with computationally intensive or high-fidelity explainability techniques.

%\paragraph{Hierarchical federated learning.}
Hierarchical FL introduces intermediate aggregation layers, such as edge servers or regional coordinators, complicating both learning and explainability.
Patni and Lee~\cite{Patni2024Explainable} integrate XAI into hierarchical FL to explain resource management and communication efficiency, while Pervez \emph{et al.}~\cite{Towards2025Pervez} apply multi-level explainability in hierarchical medical federations.
Here, explanations may exist at client, edge, and global levels simultaneously.
Hierarchical settings may additionally require explainability mechanisms operating across multiple federation levels simultaneously.

%\paragraph{Heterogeneity considerations.}
Across all settings, non-IID data distributions, partial client participation, and streaming or continual data significantly affect explanation stability and reliability.
Incremental and streaming scenarios, such as federated class-incremental learning~\cite{Explainable2025Carillo}, further challenge the consistency of explanations over time.

In conclusion, axis~E provides practical guidance for selecting explainability mechanisms that remain reliable under different federated deployment conditions and heterogeneity assumptions.

\subsection{Operationalizing the Taxonomy for FedXAI Design}
\label{subsec:taxonomy_operationalization}

The proposed taxonomy can be used to systematically identify suitable explainability strategies based on application requirements, federated settings, privacy constraints, and system-level objectives.

Rather than viewing explainability as a standalone post-hoc component, the taxonomy conceptualizes FedXAI as a multidimensional design space spanning explainability objectives, XAI mechanisms, integration strategies, and federated deployment conditions.

By jointly considering explainability objectives, XAI mechanisms, integration strategies, and federated settings, the taxonomy enables systematic comparison of existing approaches and supports informed design choices for new FedXAI systems.

In practice, the taxonomy may be operationalized sequentially by:
(i) identifying the primary role of explainability (axis~A),
(ii) selecting an appropriate XAI family (axis~B),
(iii) determining the explanation scope and target (axis~C),
(iv) choosing the integration level and exchanged artifacts (axis~D), and
(v) adapting the design to the target FL setting and heterogeneity assumptions (axis~E).

Table~\ref{tab:taxonomy_mapping} illustrates representative mappings of existing FedXAI methods onto the proposed taxonomy axes.

The table highlights how representative FedXAI methods differ in terms of explainability objectives, integration strategies, and federated deployment settings.

Although the taxonomy axes are conceptually orthogonal, practical FedXAI systems often exhibit recurring cross-axis correlations driven by domain constraints, privacy requirements, and communication limitations.
For example, cross-device systems often rely on lightweight post-hoc explainers, whereas cross-silo healthcare applications typically employ richer visualization-based or hybrid explainability mechanisms.

The taxonomy can further guide the selection of explainability mechanisms according to application and system requirements.
Applications with strict communication or privacy constraints may favor lightweight instance-level explainers, whereas high-stakes domains such as healthcare may require richer or privacy-preserving explanation mechanisms.
Similarly, personalization-oriented systems may benefit from attention-based or architecture-integrated explainability, whereas coordination-driven FL systems may incorporate explanation signals directly into clustering, aggregation, or client selection procedures.

A healthcare-oriented cross-silo FL system requiring clinically interpretable predictions and strong privacy guarantees may favor visualization-based or hybrid explainability approaches with client-side integration, such as Grad-CAM or secure SHAP aggregation. 
In contrast, resource-constrained cross-device IoT environments may prefer lightweight post-hoc explainers with minimal communication overhead. 
Similarly, systems emphasizing adaptive aggregation or personalization may select attention-based or explanation-driven coordination mechanisms.

Overall, the proposed taxonomy provides not only a structured view of the current FedXAI landscape, but also a practical foundation for designing future FedXAI systems under diverse architectural, privacy, and application constraints.

\begin{landscape}
\begin{table}[htbp]
\centering
\caption{Mapping of FedXAI methods onto the proposed multi-axis taxonomy.}
\label{tab:taxonomy_mapping}
\resizebox{\linewidth}{!}{
\begin{tabular}{p{1cm} p{4cm} p{4cm} p{4cm} p{5cm} p{5cm}}
\hline
\textbf{Paper} &
\textbf{Axis A} &
\textbf{Axis B} &
\textbf{Axis C} &
\textbf{Axis D} &
\textbf{Axis E} \\
\hline

~\cite{Explainable2025Carillo}
& Training diagnosis
& Visualization-based
& Training dynamics
& Hybrid; attribution statistics
& Continual, non-IID FL \\

~\cite{Corcuera2022}
& Transparency
& Interpretable-by-design
& Model structure
& Hybrid; fuzzy rules
& Cross-silo FL \\

~\cite{Explainable2025Yazdinejad}
& Security and trust
& Post-hoc explainers
& Prediction-level explanations
& Hybrid; encrypted explanation sharing
& Cross-device cybersecurity FL \\

~\cite{Designing2022Raza}
& Transparency
& Visualization-based
& Instance-level predictions
& Client-side; saliency maps
& Cross-silo healthcare FL \\

~\cite{Sarker2024Enhancing}
& Transparency / personalization
& Attention-based
& Prediction-level explanations
& Client-side; attention weights
& Cross-device smart-grid FL \\

~\cite{Oki2024Evaluation}
& Transparency
& Post-hoc explainers
& Model-level feature importance
& Server-side; SHAP summaries
& Non-IID intrusion detection FL \\

~\cite{Patni2024Explainable}
& System-level management
& Post-hoc explainers
& System behavior
& Server-side; system metrics
& Hierarchical FL \\

~\cite{Mastoi2025Explainable}
& Transparency
& Visualization-based
& Instance-level medical imaging explanations
& Client-side; saliency maps
& Cross-silo healthcare FL \\

~\cite{Explainable2025Ali}
& Aggregation and coordination
& Post-hoc explainers
& Cluster-level explanations
& Hybrid; SHAP-based clustering
& Cross-device energy forecasting FL \\

~\cite{Taheri2025Explainable}
& Personalization
& Post-hoc explainers
& Prediction-level explanations
& Client-side explanations
& Dynamic vehicular FL \\

~\cite{Kumar2025Federated}
& Transparency
& Post-hoc explainers
& Prediction-level explanations
& Client-side explanations
& Cross-silo healthcare FL \\

~\cite{Bilal2025}
& Transparency
& Post-hoc explainers
& Prediction-level explanations
& Client-side explanations
& Cross-device IoT FL \\

~\cite{Tawfik2025FedMedSecure}
& Personalization and security
& Attention-/architecture-based
& Prediction and confidence explanations
& Hybrid; attention and prototypes
& Cross-silo healthcare FL \\

~\cite{Corcuera2025Increasing}
& Transparency
& Interpretable-by-design
& Model structure
& Hybrid; symbolic rules
& Cross-silo FL \\

~\cite{Kalakoti2025}
& Security and trust
& Post-hoc explainers
& Model-level explanations
& Hybrid; secure SHAP aggregation
& Privacy-preserving FL \\

~\cite{Amato2025}
& Aggregation and coordination
& Attention-/architecture-based
& Cluster-level and model-level explanations
& Hybrid; semantic metadata
& Cross-silo healthcare FL \\

~\cite{Transparency2024Awosika}
& Transparency
& Post-hoc explainers
& Model-level and transaction-level explanations
& Server-side; SHAP summaries
& Cross-silo financial FL \\

~\cite{Towards2025Pervez}
& Transparency and diagnosis
& Visualization-based
& Multi-level medical explanations
& Hierarchical; saliency-based explanations
& Hierarchical healthcare FL \\

\hline
\end{tabular}
}

\end{table}

\end{landscape}

\section{FedXAI Methodological Approaches}
\label{sec:FedXAI_Methodological_Approaches}
Building upon the multi-axis taxonomy introduced in Section \ref{Multi-axisTaxonomy}, this section reviews and categorizes existing FedXAI methods according to how they instantiate different roles of explainability within the FL lifecycle. Rather than organizing works by specific XAI techniques, we focus on methodological patterns that reveal how explainability is integrated, exploited, and operationalized in federated environments.

\subsection{Post-hoc FedXAI: Explainability after Federated Training}
\label{subsec:posthoc-fedxai}
Post-hoc explainability represents the most prevalent methodological 
paradigm in the current 
FedXAI literature. 
In this setting, FL is performed using conventional optimization and aggregation schemes--most commonly variants of Federated Averaging--while explainability mechanisms are applied \emph{after} the training process, without directly influencing model optimization or aggregation dynamics. This approach enables the deployment of highly expressive black-box models, particularly deep neural networks, while augmenting them with interpretability mechanisms as an external layer.

A representative example of post-hoc FedXAI is provided by Kalakoti \emph{et al.}~\cite{Kalakoti2025}, who propose a FedXAI-based IDS framework in a horizontal FL (HFL) setting for IoT botnet detection. Their framework uses SHAP to explain the server model by securely aggregating SHAP values from client-side models, without sharing client data with the server.

Let $\mathcal{M}_{c_i}$ denote the trained model of client $c_i$, and let $\boldsymbol{\phi}_{c_i}(x) \in \mathbb{R}^d$ denote the SHAP values generated for a point of interest $x$. Leveraging the additive nature of SHAP values, the explanation of the server model $\mathcal{M}_s$ is obtained by aggregating explanations from the individual client models:

% equation-13
\begin{equation}
\boldsymbol{\phi}_s(x) = \frac{1}{|N|} \sum_{c_i \in N} \boldsymbol{\phi}_{c_i}(x_j),
\end{equation}
where $N$ denotes the set of participating clients.

To preserve the privacy of individual clients’ SHAP values, the authors employ a secure multi-party computation (SMPC) protocol based on secret sharing. This allows the server to compute aggregated SHAP values without knowing the individual SHAP values of each client. The authors evaluate this approximation by comparing the securely aggregated client-based explanations
%, denoted as  $\textcolor{red}{E_{\text{instance}}}$ , 
with server-based explanations,
%denoted as $\textcolor{red}{E_{\text{model}}}$, 
generated when the server has direct access to the data from all participating clients. Their results show that securely aggregated client-side explanations can approximate the feature attributions of the server model without relying on client data.

A dominant instantiation of post-hoc FedXAI involves \emph{client-side explanation generation}, where each participating client independently computes explanations for its local predictions using model-agnostic XAI techniques. This design aligns naturally with privacy-preserving requirements, as both raw data and explanation computations remain confined to the client side. Numerous studies in intrusion detection and cybersecurity adopt this paradigm. For instance, Kalakoti \emph{et al.}~\cite{Kalakoti2025} integrate SHAP, LIME, and Integrated Gradients into a federated LSTM-based IDS, enabling each IoT client to interpret detection outcomes on the client side. Similarly, Taheri \emph{et al.}~\cite{Taheri2025Explainable} and Harshitha \emph{et al.}~\cite{Harshitha2025Federated} employ post-hoc SHAP- and LIME-based explanations at the client level to justify intrusion alerts in connected vehicle and network security scenarios. These works demonstrate that client-side post-hoc explanations are relatively easy to deploy and effective for operational transparency, debugging, and local trust establishment.

Beyond cybersecurity, post-hoc client-level explainability has been extensively explored in healthcare and medical imaging. Raza \emph{et al.}~\cite{Designing2022Raza} adapt Grad-CAM for one-dimensional ECG signals within a federated transfer learning framework, allowing clinicians to interpret arrhythmia classification decisions within each participating institution. Similar visualization-based approaches are adopted in federated medical imaging systems for brain tumor classification~\cite{Mastoi2025Explainable}, skin cancer diagnosis~\cite{Gupta2025FedMed-XAI,Naz2025Privacy}, and viral disease detection~\cite{Srinivasulu2025Overcoming,Mahir2026PriFL}, where Grad-CAM and saliency maps highlight clinically relevant image regions without exposing sensitive patient data. Collectively, these studies illustrate that post-hoc client-side explanations are particularly well suited for high-stakes domains requiring instance-level transparency and human validation.

While client-side explanations enhance institution-specific interpretability, they provide limited insight into the model-level behavior of the federated model. To address this limitation, several works extend post-hoc FedXAI toward \emph{server-side or model-level explanations} by aggregating explanation artifacts rather than raw data or gradients. In this paradigm, clients compute client-side explanation vectors--typically SHAP values--which are then transmitted to the server and combined to produce model-level feature importance profiles. Bilal \emph{et al.}~\cite{Bilal2025} demonstrate this approach in an IoT intrusion detection setting, showing that aggregated SHAP and LIME explanations approximate centralized interpretations while preserving data locality. Similarly, Oki \emph{et al.}~\cite{Oki2024Evaluation} employ Kernel SHAP to analyze how FL recovers the feature relevance patterns of a non-distributed IDS, providing a model-level explanation of performance gains under FL.

In healthcare applications, Kumar \emph{et al.}~\cite{Kumar2025Federated} integrate SHAP-based model-level explanations into a federated liver disease prediction framework, enabling clinicians to assess whether the global federated model’s reasoning aligns with established medical knowledge. Comparable model-level explanation strategies are reported in federated flood prediction~\cite{Mahir2025Advanced} and smart agriculture~\cite{Tahir2025Federated}, where SHAP-based summaries are used to interpret feature relevance across geographically distributed clients. These studies highlight that explanation aggregation serves as a practical compromise between privacy preservation and system-level interpretability.

Corbucci  \emph{et al.}~\cite{Corbucci2023ExplainingBlackBoxes} further formalize explanation aggregation in horizontal server-based FL by proposing a SHAP variant for explaining black-box federated models without requiring server access to clients' training data. 
In their approach, each client builds a SHAP explainer using its private data and computes feature-attribution vectors for the instances to be explained; these client-side explanations are then averaged to approximate the explanation of the global model. 
Using the Adult and CoverType tabular datasets, the authors show that the aggregated client-based explanations closely match simulated server-based explanations obtained with full training-data access, suggesting that SHAP aggregation can offer a practical privacy--utility trade-off for post-hoc FedXAI. 
They also identify important directions for future work, including evaluation under non-IID data, larger client populations, peer-to-peer FL, and stronger privacy-preserving mechanisms.

A more principled instantiation of post-hoc FedXAI is introduced by Ducange \emph{et al.}~\cite{FederatedSHAP}, who propose the concept of \emph{Federated SHAP} (FedSHAP) to generate privacy-preserving and consistent SHAP explanations in FL. 
Instead of directly aggregating client-side SHAP values, their approach focuses on the federated construction of a representative background dataset, which is a critical component for computing reliable Shapley values. 
For tabular data, they employ a Federated Fuzzy C-Means (FedFCM) clustering algorithm, where clients collaboratively compute federated cluster centroids that summarize the distributed data without sharing raw samples~\cite{FederatedSHAP}. These centroids form a compact and privacy-preserving background dataset for KernelSHAP. 
For image data, they introduce a Federated GAN (FedGAN) to synthesize realistic background images, enabling GradientSHAP explanations while preventing exposure of sensitive visual data~\cite{FederatedSHAP}.

This methodology satisfies three key desiderata of FedXAI: 
(i) \emph{privacy preservation}, since only aggregated statistics or model updates are exchanged; 
(ii) \emph{explanation consistency}, as all entities rely on a common federated background dataset; and 
(iii) \emph{accuracy}, since the generated explanations closely approximate those obtained in a centralized setting. 
Compared to simple client-side SHAP aggregation schemes, FedSHAP provides a more theoretically grounded, modality-aware, and systematic framework for post-hoc explainability in federated environments.

However, transmitting explanation artifacts introduces new privacy and security considerations, as explanations may encode sensitive information about local data distributions. To mitigate this risk, several works incorporate secure aggregation mechanisms for explanations. Kalakoti \emph{et al.}~\cite{Kalakoti2025} employ Secure Multiparty Computation (SMPC) to aggregate client-side SHAP values, enabling the server to approximate model-level explanations without direct access to local explanation vectors. In a complementary direction, Yazdinejad \emph{et al.}~\cite{Explainable2025Yazdinejad} combine SHAP with homomorphic encryption and differential privacy in a federated threat detection framework, explicitly addressing confidentiality risks associated with both model updates and explanation sharing.

Despite their practical appeal, post-hoc FedXAI approaches exhibit inherent limitations. Since explainability is decoupled from the federated optimization process, explanations remain passive diagnostic tools that cannot influence convergence behavior, aggregation robustness, or personalization. Moreover, explanation inconsistency across clients and across federated rounds--exacerbated by non-IID data distributions--poses challenges for trustworthiness and regulatory validation, as highlighted in comparative healthcare studies~\cite{Ducange2024Federated}. These limitations motivate more tightly integrated FedXAI methodologies, in which explainability actively shapes learning and coordination mechanisms, as discussed in the following subsections.

\subsection{Explainability-aware FL: XAI inside the FL Loop}
\label{subsec:explainability-aware-fedxai}

While post-hoc FedXAI frameworks treat explainability as an external interpretability layer, a growing body of work integrates XAI directly into the FL process, allowing explanation signals to actively influence learning dynamics, aggregation strategies, personalization, and system-level coordination. In these \emph{explainability-aware} approaches, explainability is embedded \emph{inside} the FL loop, shaping how models are trained, aggregated, and deployed rather than merely explaining their outputs.

%\paragraph{Explanation-guided aggregation.}
A first and prominent methodological pattern in explainability-aware FedXAI is the use of explanation signals to guide aggregation. Instead of relying solely on data volume or local loss for weighting client updates, these methods exploit feature attribution scores or explanation-derived relevance measures to modulate aggregation behavior.
Saleem \emph{et al.}~\cite{Weighted2025Saleem} propose a Weighted Explainable Federated Learning (WFL-XAI) framework for privacy-preserving and scalable energy optimization in autonomous vehicular networks. The framework integrates XAI techniques, including SHAP and LIME, to enhance transparency and interpretability in AI-driven energy optimization decisions. In addition, the framework introduces a weighted federated aggregation mechanism in which client contributions are dynamically adjusted according to local model performance and client data volume, enabling a more accurate and fair federated model construction in heterogeneous vehicular environments.

For the aggregation layer, clients send local model parameters and performance metrics to the central server. The client contribution weight is computed as

%equation=7,8
\begin{equation}
Q_k = \frac{R_k^2 \cdot n_k}{\sum_{i=1}^{K} R_i^2 \cdot n_i},
\label{eq:wfl_weight_Weighted2025Saleem}
\end{equation}
where $Q_k$ is the aggregation weight for client $k$, $R_k^2$ denotes the coefficient of determination of client $k$, $n_k$ is the number of samples at client $k$, and $K$ is the total number of participating clients.

The global model is then aggregated as
\begin{equation}
W^{*} = \sum_{k=1}^{K} Q_k \theta_k,
\label{eq:wfl_agg_Weighted2025Saleem}
\end{equation}
where $W^{*}$ denotes the global weight matrix and $\theta_k$ represents the local model parameters of client $k$. This weighted aggregation mechanism ensures that higher-quality local models exert greater influence on the global model.

A related explanation-guided aggregation strategy is employed in the smart grid and solar energy forecasting domain. Ali \emph{et al.}~\cite{Explainable2025Ali} propose Explainable Clustered Federated Learning (XCFL), which integrates SHAP-based feature contribution analysis into CFL for solar power forecasting. The framework improves both model performance and interpretability by incorporating feature-level contributions into cluster and global aggregation processes. In contrast to conventional FedAvg, XCFL performs weighted aggregation using feature contribution scores extracted from XAI tools, allowing more informative features and clusters to contribute more significantly to the global model.

%\paragraph{Explanation-driven clustering and heterogeneity management.}
Beyond direct aggregation weighting, explainability is also utilized for client clustering under heterogeneous data distributions. In XCFL~\cite{Explainable2025Ali}, photovoltaic clients are grouped using Mean-Shift clustering based on similar data distributions and feature characteristics. The clustered training process enables localized models to adapt to different climatic, geographic, and sensor conditions, while SHAP explanations provide interpretable insights into feature contributions at instance-level, cluster-level, and model-level. The authors show that explanation-driven clustering improves robustness and generalization under non-IID settings.

In XCFL~\cite{Explainable2025Ali}, clients are first grouped into clusters and local models are trained independently inside each cluster. SHAP-based feature contribution scores are then used to perform hierarchical weighted aggregation at both cluster and global levels. The cluster-level aggregation is formulated as
\begin{equation}
\omega_{m}^{(t+1)}
=
\sum_{k \in C_m}
\sum_{f=1}^{F}
\left(
\frac{\delta_f^{(k)}}
{\sum_{k' \in C_m}\delta_f^{(k')}}
\right)
\left(
1-\sum_{\substack{j=1 \\ j\neq f}}^{F} a_j
\right)
\omega_k^{(t)},
\label{eq:xcfl_cluster_agg}
\end{equation}
where $m$ denotes the cluster index, $C_m$ is the set of clients belonging to cluster $m$, $\delta_f^{(k)}$ denotes the SHAP-based contribution score of feature $f$ for client $k$, $a_j$ is the weight assigned to feature $j$, and $\omega_k^{(t)}$ represents the local model parameters of client $k$ at communication round $t$.

The server subsequently performs global aggregation across clusters as
\begin{equation}
\omega^{t+1}
=
\sum_{k \in C_m}
\sum_{f=1}^{F}
\frac{\delta_f^{(k)}}
{\sum_{m \in C_m}\delta_{fm}}
\left(
1-\sum_{\substack{j=1 \\ j\neq i}}^{F} a_j
\right)
\omega_m^{t},
\label{eq:xcfl_global_agg}
\end{equation}
where $\omega_m^{(t)}$ denotes the cluster-level model parameters and $\delta_{fm}$ represents the aggregated feature contribution score for feature $f$ within cluster $m$. This hierarchical SHAP-guided aggregation mechanism allows XCFL to prioritize informative features and clusters during model fusion, thereby improving forecasting accuracy, interpretability, and robustness in heterogeneous federated environments.

%\paragraph{Explainability-aware personalization and robustness.}
Several explainability-aware FedXAI frameworks integrate explanation analysis to support adaptive personalization and robustness against unreliable or malicious clients. In connected vehicular networks, Taheri \emph{et al.}~\cite{Taheri2025Explainable} combine SHAP- and LIME-based explanations with robust aggregation mechanisms to enhance both interpretability and resilience to adversarial behavior. Explanation patterns are analyzed alongside prediction outcomes to identify influential features and adjust aggregation behavior, enabling adaptive trust management across vehicles.

Similarly, FLEX-IDS~\cite{Chowdhury2026FLEX-IDS} incorporates post-hoc explanations into an explainability-aware federated IDS framework that dynamically selects among multiple aggregation strategies (e.g., FedAvg, FedProx, FedNova, FedPer). Although explanations do not directly modify gradients, they inform robustness analysis and client influence control, illustrating a hybrid form of explainability-aware personalization where XAI supports adaptive decision-making within the federated loop.

%\paragraph{Explainability-driven system coordination in hierarchical FL.}
Explainability-aware methodologies also extend beyond model-centric decisions to system-level coordination, particularly in Hierarchical FL (HFL). In such architectures, explainability is used to interpret and guide orchestration decisions rather than model predictions.

Patni and Lee~\cite{Patni2024Explainable} propose an explainable AI-empowered resource management framework for HFL, where SHAP and LIME explanations are applied to resource prediction models governing communication scheduling and node selection. Explainability enables transparent justification of system-level decisions across cloud-edge-device hierarchies, improving communication efficiency and convergence reliability. A complementary system-level integration is observed in the FLaaS-based architecture proposed by Corcuera Bárcena \emph{et al.}~\cite{Corcuera2023Enabling}, where explainable models are natively embedded into federated services to support governance, monitoring, and trust in beyond-5G/6G network environments.

%\paragraph{Semantic and knowledge-guided explainability-aware FL.}
A more advanced form of explainability-aware FedXAI incorporates domain knowledge into the learning and aggregation process itself. Amato and Branco~\cite{Amato2025} introduce SemFedXAI, a semantic framework that integrates ontologies and knowledge graphs into FL pipelines. In this approach, explainability directly influences aggregation through semantic weighting mechanisms, aligning feature relevance with clinically meaningful concepts. By elevating explanations from numerical attributions to concept-level reasoning, SemFedXAI mitigates non-IID effects and enhances interpretability in healthcare applications.

\subsection{Interpretable-by-Design FedXAI: FL of Transparent Models}
\label{subsec:interpretable-by-design-fedxai}

Interpretable-by-design FedXAI constitutes a distinct methodological paradigm from post-hoc and explainability-aware approaches. In this setting, explainability is not provided by auxiliary interpretation mechanisms, nor used as a control signal within the learning loop; instead, transparency is an intrinsic property of the learned model itself. The model representation--typically expressed as rules or tree-based fuzzy structures--is inherently human-understandable, enabling direct inspection of the decision logic without relying on external explainers.

A central challenge in interpretable-by-design FedXAI lies in reconciling symbolic or rule-based model representations with FL protocols originally designed for gradient-based optimization. As a result, these approaches often require bespoke training and aggregation mechanisms that differ substantially from standard FedAvg-style procedures.

%\paragraph{FL of fuzzy rule-based systems.}
Fuzzy Rule-Based Systems (FRBSs), particularly Takagi-Sugeno-Kang Fuzzy Rule-Based Systems (TSK-FRBSs), represent one of the most widely studied inherently interpretable models in federated settings. Corcuera Bárcena \emph{et al.}~\cite{Corcuera2022} propose a federated approach for learning explainable TSK-FRBSs in regression tasks. In their framework, each client independently learns a local TSK-FRBS from private data and transmits the learned fuzzy rules to a central server for aggregation. Unlike FedAvg-based approaches that aggregate gradients or model tensors, the server aggregates rule-based models by resolving conflicts among rules with identical antecedents and different consequents, while preserving the interpretability of the global TSK-FRBS.

In~\cite{Corcuera2022}, each fuzzy rule is assigned a rule weight $RW_k$ based on its fuzzy confidence $Conf_k$ and fuzzy support $Supp_k$:
\begin{equation}
RW_k=
2\times
\frac{Supp_k \times Conf_k}
{Supp_k+Conf_k}.
\end{equation}

At the server side, conflicting rules, i.e., rules with the same antecedent but different consequents, are resolved through weighted aggregation. Let $CR$ be the set of conflicting rules, and let $\Gamma_l$ and $rw_l$ denote the consequent coefficient vector and rule weight of the $l$-th rule in $CR$. The merged consequent is computed as
\begin{equation}
\Gamma=
\frac{
\sum_{l=1}^{|CR|}
\Gamma_l \cdot rw_l
}{
\sum_{l=1}^{|CR|}
rw_l
},
\end{equation}

and the merged rule weight is
\begin{equation}
rw=
\frac{1}{|CR|}
\sum_{l=1}^{|CR|}
rw_l.
\end{equation}

This aggregation strategy preserves the linguistic structure and interpretability of the global TSK-FRBS while avoiding raw data sharing among clients.

Subsequent works by Corcuera Bárcena \emph{et al.}~\cite{Corcuera2023Enabling,Corcuera2025Increasing} further extend this idea by embedding federated fuzzy models into system-level architectures, including FL as-a-Service (FLaaS) frameworks for beyond-5G/6G networks. These studies show that interpretable fuzzy models can be trained federatively while maintaining human-readable fuzzy rules and achieving performance comparable to centralized learning.

%\paragraph{Federated induction of fuzzy regression trees.}
Fuzzy Regression Trees (FRTs) extend the interpretable-by-design paradigm by organizing decision logic hierarchically, enabling implicit feature selection and improved scalability. In~\cite{Corcuera2025Increasing}, Corcuera Bárcena \emph{et al.} introduce an FL algorithm for FRTs that relies on sharing aggregated sufficient statistics--such as weighted sums of squares--rather than raw samples or gradients. The server uses these statistics to construct a global fuzzy tree through a federated tree induction procedure analogous to centralized training.

This statistics-based federation mechanism highlights a key insight of interpretable-by-design FedXAI: privacy preservation and interpretability can be jointly achieved by carefully designing what information is exchanged during federation. By avoiding iterative gradient exchange, federated FRTs maintain transparency while achieving competitive predictive performance.

%\paragraph{Comparative studies and hybrid interpretability.}
Several works explicitly compare interpretable-by-design models with black-box models augmented by post-hoc explainability under identical federated conditions. Ducange \emph{et al.}~\cite{Ducange2024Federated} conduct a comprehensive healthcare case study on Parkinson’s disease progression prediction, contrasting federated TSK-FRBSs with federated multilayer perceptrons explained via SHAP. While the neural model achieves slightly higher predictive accuracy, the interpretable-by-design approach exhibits superior transparency, explanation consistency across clients, and ease of clinical interpretation. This comparative analysis underscores that trustworthiness in FedXAI cannot be assessed through accuracy alone, but must also account for model-level interpretability interpretability and stability.

%\paragraph{Tooling and infrastructure for interpretable-by-design FedXAI.}
Beyond algorithmic contributions, practical adoption of interpretable-by-design FedXAI depends on appropriate tooling and software abstractions. Daole \emph{et al.}~\cite{OpenFL2023} address this gap through \emph{OpenFL-XAI}, an extension of the Intel OpenFL framework that enables FL of rule-based and fuzzy models. Rather than exchanging neural network weights, OpenFL-XAI supports the transmission and aggregation of symbolic rule representations encoded as tensors. This contribution demonstrates that explainability in FL is not solely a modeling concern, but also an infrastructural challenge requiring tailored orchestration and aggregation logic.

%\paragraph{System-level interpretable-by-design FedXAI.}
Interpretable-by-design models have also been integrated into system-level federated architectures. In FLaaS environments for beyond-5G/6G networks~\cite{Corcuera2023Enabling}, inherently explainable fuzzy models are deployed as shared services, enabling transparency, accountability, and trust in distributed AI-driven network management. In these settings, interpretability supports not only human understanding but also service governance and operational monitoring across heterogeneous stakeholders.

%\subsection{Summary and Discussion}
%\label{subsec:fedxai-summary-discussion}

This section has reviewed FedXAI methodologies through a role-oriented lens, focusing on \emph{how} explainability is instantiated and operationalized within the FL lifecycle rather than on specific XAI techniques. The surveyed literature reveals a clear methodological progression from loosely coupled post-hoc explainability toward tightly integrated and intrinsically transparent FL paradigms.

Post-hoc FedXAI approaches, reviewed in Section~\ref{subsec:posthoc-fedxai}, dominate current applications due to their architectural simplicity and compatibility with existing FL frameworks. By decoupling learning from explanation, these methods enable the deployment of expressive black-box models while satisfying basic transparency requirements through client-side or aggregated explanations. This paradigm has proven particularly effective in high-stakes domains such as healthcare, cybersecurity, and finance, where instance-level explanations support human validation and regulatory compliance. However, the passive nature of post-hoc explanations limits their influence on learning dynamics, robustness, and personalization. Moreover, explanation instability under non-IID data distributions and the potential privacy leakage associated with explanation sharing pose fundamental challenges to trustworthiness and governance.

Explainability-aware FedXAI methods, discussed in Section~\ref{subsec:explainability-aware-fedxai}, represent a conceptual shift in which explainability evolves from a diagnostic artifact into an active control signal within the FL loop. By leveraging explanation signals to guide aggregation, clustering, personalization, and system-level coordination, these approaches explicitly acknowledge that transparency can improve not only interpretability but also learning efficiency and robustness. Explanation-guided weighting, clustering, and resource management demonstrate that explanation vectors can serve as high-level semantic representations of client data, offering principled mechanisms for managing heterogeneity and uncertainty. Nevertheless, this tighter integration introduces additional complexity, computational overhead, and new attack surfaces, particularly related to the integrity and privacy of explanation signals themselves.

Interpretable-by-design FedXAI approaches, reviewed in Section~\ref{subsec:interpretable-by-design-fedxai}, constitute the strongest and most principled form of federated explainability. By embedding transparency directly into the model structure--through fuzzy rule-based systems, fuzzy regression trees, and other symbolic representations--these methods eliminate reliance on post-hoc explainers and provide globally consistent, human-readable decision logic. Crucially, they demonstrate that privacy preservation and explainability can be jointly achieved through careful redesign of federated training and aggregation protocols, often by exchanging aggregated statistics or symbolic parameters rather than gradients. While interpretable-by-design approaches offer superior explanation consistency and auditability, their applicability is currently constrained to structured-data domains and requires bespoke federated infrastructures and aggregation mechanisms.

Taken together, the three methodological paradigms reveal an inherent trade-off between \emph{model expressiveness}, \emph{system complexity}, and \emph{explainability guarantees}. Post-hoc methods favor flexibility and ease of deployment, explainability-aware approaches balance transparency with adaptive coordination, and interpretable-by-design models prioritize trust and governance at the cost of reduced expressiveness and increased system specialization. Importantly, these paradigms should not be viewed as mutually exclusive. Several recent works suggest hybrid directions that combine interpretable cores with explainability-aware coordination or selectively apply post-hoc explanations atop intrinsically transparent models.

From a broader perspective, the evolution of FedXAI methodologies reflects a shift from viewing explainability as an optional add-on toward recognizing it as a first-class design constraint in distributed learning systems. As FL continues to expand into regulated, safety-critical, and large-scale environments, future FedXAI research must address open challenges related to explanation consistency, privacy-preserving explanation aggregation, robustness against explanation manipulation, and standardized evaluation protocols. Ultimately, achieving trustworthy federated AI will require principled integration of explainability across algorithmic, system, and governance layers rather than isolated methodological advances.

\section{Evaluation of FedXAI}
\label{sec:evaluation-fedxai}

Despite the rapid growth of FedXAI, its evaluation remains largely ad hoc and fragmented. Most existing studies primarily report predictive performance and provide qualitative explanation visualizations, while lacking a unified and principled evaluation methodology tailored to federated explainability. As highlighted in early conceptual works on FedXAI~\cite{Corcuera2022Fed-XAI}, the absence of standardized evaluation protocols constitutes a fundamental research gap that limits reproducibility, comparability, and real-world trustworthiness. This section formalizes the \emph{FedXAI evaluation gap} by clarifying what should be evaluated, how evaluation is currently conducted, and why existing benchmarks are insufficient.

\subsection{What Should Be Evaluated in FedXAI?}
\label{subsec:fedxai-what-to-evaluate}

Evaluation in FedXAI must extend beyond classical performance metrics and explicitly account for explainability, privacy, robustness, and system-level costs induced by federation.

%\paragraph{Predictive performance.}
Most FedXAI studies continue to assess predictive performance using task-specific metrics such as accuracy, F1-score, RMSE, or AUC, comparing centralized, local-only, and federated settings~\cite{Explainable2025Carillo,Mahir2025Advanced,Demystifying2024Patel,Sarker2024Enhancing,Mastoi2025Explainable}. While necessary, predictive performance alone is insufficient, as models with similar accuracy may exhibit substantially different explanation behaviors across clients and federated rounds~\cite{Oki2024Evaluation,Kalakoti2025}.

%\paragraph{Explanation quality.}
A defining goal of FedXAI is to generate explanations that remain meaningful under data decentralization and heterogeneity. Consequently, explanation quality must be evaluated explicitly, typically along dimensions such as fidelity, stability, sparsity, and consistency. Several works assess explanation fidelity using attribution stability or insertion--deletion analysis in image and signal domains~\cite{Choir2016Sahoo,Towards2025Pervez}, while others report cross-round or cross-client consistency of SHAP- or Grad-CAM-based explanations~\cite{Explainable2025Carillo,Explainable2025Ali,Ducange2024Federated}. However, these metrics are rarely standardized, and their sensitivity to non-IID data distributions remains an open challenge.

%\paragraph{Privacy leakage from explanations.}
Although FL prevents raw data sharing, explanations themselves may leak sensitive information about local data distributions or rare features. This risk is explicitly acknowledged in privacy-preserving FedXAI frameworks that combine explainability with differential privacy or secure aggregation~\cite{Explainable2025Yazdinejad,Tawfik2025FedMedSecure,Kalakoti2025}. Nevertheless, systematic quantification of explanation-induced privacy leakage is still largely absent from FedXAI evaluation practices.

%\paragraph{Robustness and security.}
Federated systems are vulnerable to adversarial behaviors such as poisoning, backdoor insertion, and malicious client updates. In FedXAI, robustness evaluation must therefore consider not only predictive degradation under attack, but also manipulation or distortion of explanations. Recent intrusion-detection studies highlight the need to evaluate explanation reliability under adversarial participation~\cite{Taheri2025Explainable,Chowdhury2026FLEX-IDS}, yet explanation-aware attack models remain underexplored.

%\paragraph{Computational and communication cost.}
Explainability introduces additional overhead in federated systems, including client-side explanation computation, explanation aggregation, and increased communication payloads. Several works explicitly report communication and computational overhead associated with SHAP, Grad-CAM, or ensemble-based explainability~\cite{Choir2016Sahoo,Patni2024Explainable,Bilal2025}. However, cost metrics are rarely analyzed jointly with explanation quality, obscuring practical trade-offs in resource-constrained deployments.

\subsection{Metrics and Evaluation Protocols}
\label{subsec:fedxai-metrics-protocols}

Current FedXAI evaluations rely on heterogeneous metrics drawn from centralized XAI, FL, and application-specific traditions, limiting cross-study comparability.

%\paragraph{Explanation stability across rounds and clients.}
Stability metrics quantify how explanations evolve across federated rounds or differ between clients. Several studies measure distances between feature-attribution vectors (e.g., SHAP value distributions) to assess explanation drift under non-IID data~\cite{Oki2024Evaluation,Harshitha2025Federated,Bilal2025}. Stability is particularly critical in cross-silo and safety-critical domains, where inconsistent explanations may erode trust.

%\paragraph{Explanation agreement.}
Agreement metrics assess consistency between different explanation methods (e.g., SHAP vs.\ LIME), between clients, or between instance-level and model-level explanations. Such analyses are used to justify explanation reliability in intrusion detection, healthcare, and energy systems~\cite{Choir2016Sahoo,Taheri2025Explainable,Bilal2025}. Low agreement may indicate either model uncertainty or explanation instability induced by federation.

%\paragraph{Fidelity and insertion--deletion metrics.}
For image-based and signal-based tasks, explanation fidelity is often evaluated using insertion and deletion curves, which quantify the impact of salient regions on model predictions~\cite{Towards2025Pervez}. 
While widely adopted, these metrics were originally designed for centralized models and may not fully capture the specific dynamics of federated explanation settings. 

A comprehensive evaluation of explanation quality in FL is provided by Ducange \emph{et al.}~\cite{FederatedSHAP}, who explicitly quantify how closely federated SHAP explanations approximate their centralized counterparts. 
They measure explanation accuracy by computing the Frobenius norm between the SHAP attribution matrices obtained under different background generation strategies and those produced using the full centralized training data.

This metric directly captures the discrepancy between federated and centralized explanations, offering a principled and quantitative way to assess explanation fidelity in federated settings. 
Their results show that Federated SHAP significantly reduces the explanation error compared to random or locally perturbed background datasets, sometimes by up to a factor of two or three, depending on the data modality.

Moreover, they demonstrate that explanation quality remains stable even when using compact synthetic backgrounds (e.g., a few thousand GAN-generated images), highlighting an effective trade-off between computational efficiency and explanation accuracy. 
This work thus establishes a concrete quantitative benchmark for evaluating explanation quality in FedXAI and represents one of the most rigorous validation frameworks currently available in the literature.

%\paragraph{Calibration-aware explainability.}
A smaller subset of works explicitly consider calibration alongside explainability, arguing that well-calibrated predictions with unstable explanations remain unsafe in high-stakes applications~\cite{Choir2016Sahoo,Kumar2025Federated}. Joint evaluation of calibration and explanation quality represents a promising yet underdeveloped direction in FedXAI.

%\paragraph{FedXAI-specific metrics.}
Some studies introduce metrics tailored to federated explainability, such as cross-client explanation distance, explanation entropy, semantic consistency, or convergence of explanations across rounds~\cite{Explainable2025Ali,Ducange2024Federated,Amato2025}. However, these metrics are rarely reused across papers, preventing the emergence of a shared evaluation standard.
Table~\ref{tab:fedxai-eval-core} summarizes a core evaluation suite for FedXAI, highlighting the key dimensions, commonly used metrics, and representative studies.

\begin{table}[t]
\centering
\caption{A core evaluation suite for FedXAI}%%, summarizing key evaluation aspects, representative metrics, and example studies.
\label{tab:fedxai-eval-core}
\small
\begin{tabular}{p{3cm} p{6.5cm} p{3cm}}
\hline
\textbf{Evaluation Aspect} & \textbf{Metrics / Methods} & \textbf{Representative Works} \\
\hline
Predictive Performance &
Accuracy, F1-score, RMSE, AUC &
\cite{Explainable2025Carillo,Mahir2025Advanced,Demystifying2024Patel,Sarker2024Enhancing} \\
\hline
Explanation Quality &
Fidelity, sparsity, insertion--deletion, calibration, discrepancy to centralized explanations (e.g., Frobenius norm on SHAP matrices) &
\cite{Kumar2025Federated,Choir2016Sahoo,Towards2025Pervez,FederatedSHAP} \\

\hline
Explanation Stability &
Cross-round / cross-client attribution distance &
\cite{Oki2024Evaluation,Explainable2025Ali,Bilal2025} \\
\hline
Privacy Risk &
Differential privacy, secure aggregation, leakage analysis &
\cite{Explainable2025Yazdinejad,Tawfik2025FedMedSecure,Kalakoti2025} \\
\hline
Robustness &
Poisoning/backdoor resilience, explanation integrity &
\cite{Taheri2025Explainable,Chowdhury2026FLEX-IDS} \\
\hline
System Cost &
Communication overhead, computation latency &
\cite{Choir2016Sahoo,Patni2024Explainable,Bilal2025} \\
\hline
\end{tabular}
\end{table}

\subsection{The Benchmarking Gap in FedXAI}
\label{subsec:fedxai-benchmarking-gap}

Despite extensive empirical experimentation, the FedXAI literature lacks dedicated benchmarks explicitly designed for federated explainability. Most studies rely on centralized public datasets that are artificially partitioned to simulate federated clients~\cite{Sarker2024Enhancing,Mastoi2025Explainable,Fatema2025Federated,Naz2025Privacy}. While this approach improves reproducibility, it fails to capture real-world federation dynamics such as evolving data streams, institutional biases, asynchronous participation, and explanation leakage risks.

Moreover, most datasets do not provide ground-truth explanations, forcing researchers to rely on proxy metrics such as stability, sparsity, or fidelity without validating whether explanations align with domain knowledge~\cite{Explainable2025Ali, Choir2016Sahoo,Towards2025Pervez}. As emphasized in survey and conceptual works~\cite{Corcuera2022Fed-XAI}, this limitation hinders systematic comparison and obscures whether reported explanations genuinely support human understanding and decision-making.

These observations highlight the urgent need for \emph{federated explainability benchmarks} that jointly define realistic client heterogeneity, standardized evaluation protocols, privacy-aware explanation constraints, and domain-relevant explanation targets. Without such benchmarks, progress in FedXAI evaluation will remain fragmented, slowing adoption in regulated and safety-critical domains.

\section{Datasets and Data Characteristics}
\label{subsec:datasets-fedxai}
Dataset characteristics in FedXAI differ fundamentally from those in centralized XAI benchmarks, where data homogeneity and unified annotation schemas are often implicitly assumed.

In FedXAI, data are not only distributed across clients, but also intrinsically heterogeneous, privacy-constrained, and often collected under distinct operational or institutional contexts. As a result, dataset characteristics directly influence the design of FL protocols, the choice of explainability techniques, and the interpretation of generated explanations. Rather than being a neutral input, data properties in FedXAI actively shape what types of explanations are feasible, stable, and meaningful.

%%\paragraph{Data modality and its impact on explainability.}
FedXAI has been applied across a wide range of data modalities, including medical imaging, physiological time-series, tabular records, network traffic, and multimodal sensor data. Each modality affords different forms of explainability and constrains how explanations can be communicated to human stakeholders. In medical imaging applications, such as brain tumor classification~\cite{Mastoi2025Explainable}, skin cancer detection~\cite{Gupta2025FedMed-XAI,Naz2025Privacy}, eye disease diagnosis~\cite{KbFL2025mahadi}, and leukemia diagnosis~\cite{Towards2025Pervez}, datasets consist of high-dimensional images with strong spatial structure. Consequently, explainability is predominantly visual and region-based, relying on saliency maps, Grad-CAM, occlusion sensitivity, or attention heatmaps to highlight diagnostically relevant areas. In these settings, explanations are tightly coupled to spatial coherence and clinical plausibility rather than feature ranking.

In contrast, tabular and network-centric datasets used in cybersecurity and finance--such as encrypted traffic classification~\cite{Explainable2025Carillo}, IoT intrusion detection~\cite{Fatema2025Federated,Bilal2025,Kalakoti2025}, CPSS threat detection~\cite{Explainable2025Yazdinejad}, and financial fraud detection~\cite{Aljunaid2025Secure,Transparency2024Awosika}--naturally lend themselves to feature-attribution explanations. Here, explainability is expressed through ranked feature contributions (e.g., packet statistics, protocol fields, transaction attributes), typically generated via SHAP or LIME. The dataset structure enables both instance-level explanations (per alert or transaction) and model-level summaries that support auditing and policy compliance. Time-series datasets, including ECG monitoring~\cite{Designing2022Raza,Manocha2025Federated} and smart grid load forecasting~\cite{Sarker2024Enhancing}, occupy an intermediate position, where explanations often highlight influential temporal segments or lagged variables rather than static features or spatial regions.

%%\paragraph{Client heterogeneity and sources of non-IID data.}
A defining property of datasets in FedXAI is statistical heterogeneity across clients, commonly referred to as non-IID data. This heterogeneity arises from multiple sources, including institutional differences (e.g., hospitals using different imaging devices or protocols~\cite{Ducange2024Federated, Mastoi2025Explainable}), geographical and environmental variation (e.g., solar farms and smart grids under different climate conditions~\cite{Explainable2025Ali,Sarker2024Enhancing}), device-level diversity in IoT and vehicular networks~\cite{Choir2016Sahoo,Weighted2025Saleem}, and severe class imbalance in rare-event detection tasks such as fraud and intrusion detection~\cite{Aljunaid2025Secure, Fatema2025Federated}. Importantly, non-IID data affect not only predictive performance but also the stability and consistency of explanations.

Several studies explicitly demonstrate that explanation patterns vary across clients under heterogeneous data distributions. For example, in federated intrusion detection, SHAP-based feature importance profiles differ significantly between client-side and federated models when data are skewed~\cite{Oki2024Evaluation,Kalakoti2025}. Similarly, in healthcare applications, client-specific data characteristics lead to variations in saliency maps or feature attributions, raising concerns about whether a single model-level explanation accurately reflects local decision logic~\cite{Kumar2025Federated, Ducange2024Federated}. These observations highlight that dataset heterogeneity is an explainability challenge in its own right, motivating the use of clustered FL~\cite{Explainable2025Ali}, weighted aggregation strategies~\cite{Weighted2025Saleem}, or semantic enrichment~\cite{Amato2025} to mitigate explanation divergence.

%%\paragraph{Simulated federations and dataset realism.}
Despite the practical motivation of FedXAI, a substantial portion of the literature relies on simulated federated settings constructed by partitioning centralized public datasets into artificial clients. This approach is common in healthcare~\cite{Kumar2025Federated, Mastoi2025Explainable,Naz2025Privacy}, cybersecurity~\cite{Bilal2025, Fatema2025Federated}, energy forecasting~\cite{Explainable2025Ali, Sarker2024Enhancing}, and finance~\cite{Transparency2024Awosika, Aljunaid2025Secure}. While simulation enables reproducibility and controlled experimentation, it often fails to capture real-world federation dynamics such as evolving data streams, cross-institutional biases, asynchronous participation, and concept drift. As noted in several studies, explanation quality observed under simulated conditions may overestimate the robustness and consistency achievable in operational deployments~\cite{Oki2024Evaluation,Kalakoti2025}.

Another limitation concerns the lack of ground-truth explanations in most datasets. Except for a few medical imaging tasks where expert-annotated regions are implicitly available, the majority of FedXAI datasets do not provide explicit rationales against which explanations can be quantitatively validated. Consequently, many works evaluate explainability qualitatively or through proxy metrics such as stability, sparsity, or fidelity~\cite{Explainable2025Ali, Choir2016Sahoo,Towards2025Pervez}. While these metrics are informative, they do not fully address whether explanations align with domain knowledge or support correct human decision-making, underscoring a key gap in current FedXAI benchmarks.

%%\paragraph{Implications for FedXAI design and evaluation.}
The reviewed literature indicates that dataset characteristics are inseparable from FedXAI methodology. Data modality constrains the choice of explainability techniques, client heterogeneity affects explanation stability and aggregation, and dataset realism determines the external validity of reported interpretability claims. As a result, rigorous FedXAI evaluation requires explicit documentation of dataset properties, including modality, client partitioning strategy, degree and source of non-IID distributions, and availability of explanation-relevant annotations. Failure to account for these factors risks conflating algorithmic explainability with dataset-specific artifacts.

To facilitate systematic comparison across studies, Table~\ref{tab:fedxai-datasets} summarizes representative dataset characteristics used in FedXAI research, organized by application domain, data modality, heterogeneity type, and explainability implications. This structured overview highlights recurring patterns and open challenges in dataset design and selection for trustworthy and interpretable federated AI systems.

\begin{landscape}
\begin{table}[p]
\centering
\caption{Representative datasets and data characteristics used in FedXAI studies across application domains.}
\label{tab:fedxai-datasets}

\renewcommand{\arraystretch}{1.2}
\setlength{\tabcolsep}{6pt}
\footnotesize

\begin{tabular}{p{4cm} p{1.5cm} p{5cm} p{2.4cm} p{3.0cm} p{3.6cm}}
\hline
\textbf{Domain} &
\textbf{Modality} &
\textbf{Datasets} &
\textbf{Clients} &
\textbf{non-IID Sources} &
\textbf{Explainability Implications} \\
\hline

Healthcare (Imaging) &
Images &
Brain Tumor MRI~\cite{Mastoi2025Explainable}; HAM10000~\cite{Gupta2025FedMed-XAI,Naz2025Privacy} &
Hospitals &
Class imbalance; acquisition bias &
Spatial explanations (Grad-CAM, saliency) \\
\hline

Healthcare (Signals) &
Time-series &
MIT-BIH~\cite{Designing2022Raza,Manocha2025Federated} &
Edge devices &
Noise; sampling heterogeneity &
Temporal segment attribution \\
\hline

Cybersecurity / IDS &
Tabular &
CICIDS2017; N-BaIoT~\cite{Bilal2025, Fatema2025Federated,Explainable2025Carillo} &
IoT gateways &
Protocol skew; imbalance &
Feature attribution (SHAP/LIME) \\
\hline

Energy / Smart Grid &
Time-series &
Panama Load~\cite{Sarker2024Enhancing}; German Solar~\cite{Explainable2025Ali} &
Grid nodes &
Geographical variability &
Cluster-level explanations \\
\hline

Vehicular Networks &
Telemetry &
Urban traffic~\cite{Weighted2025Saleem} &
Vehicles &
Dynamic context &
Explanation-guided weighting \\
\hline

Finance &
Tabular &
Bank fraud~\cite{Transparency2024Awosika, Aljunaid2025Secure} &
Banks &
Extreme imbalance &
Regulatory-compliant explanations \\
\hline

\end{tabular}
\end{table}
\end{landscape}

\section{Application Domains and Use Cases}
\subsection{Healthcare and Medical Imaging}

Healthcare is one of the most prominent application domains of FedXAI, driven by strict data privacy regulations and the need for clinically meaningful explanations~\cite{Ducange2026FDTNuclei}. A representative line of work focuses on ECG-based cardiac monitoring, where FL enables collaborative arrhythmia classification without centralizing sensitive patient data, while explainability supports clinical interpretability. Studies such as Raza \emph{et al.}~\cite{Designing2022Raza} and Manocha \emph{et al.}~\cite{Manocha2025Federated} demonstrate how post-hoc attribution methods, adapted to one-dimensional signals, highlight salient temporal segments of ECG recordings, allowing clinicians to assess the physiological relevance of model predictions. These works illustrate a recurring pattern in healthcare FedXAI, where explainability primarily serves to justify predictions at the signal level rather than to audit model behavior.

Recent advances further extend FedXAI toward multimodal and privacy-preserving clinical decision support systems. 
Qazi \emph{et al.}~\cite{Qazi2025FDL} propose a federated deep learning framework integrating Vision Transformers (ViT), DINOv2-based self-supervised learning, and FedProx to support personalized medical image analysis under heterogeneous and non-IID healthcare environments. 
Their framework combines differential privacy and elliptic curve cryptography (ECC) for secure federated communication, while Grad-CAM and LIME provide instance-level explainability for clinical transparency across tuberculosis, diabetic retinopathy, and brain tumor diagnosis tasks.

Chorney and Wang~\cite{Chorney2024FTL} investigate federated transfer learning for ECG analysis under realistic clinical settings with heterogeneous and non-IID data distributed across institutions. 
Their framework employs autoencoders to map ECG signals with different dimensions and lead configurations into a shared latent space, enabling privacy-preserving federated training across heterogeneous datasets. 
The study highlights the practical challenges of deploying federated healthcare models beyond idealized centralized assumptions and demonstrates the importance of handling heterogeneous medical data in realistic FedXAI environments.

Beyond time-series analysis, medical imaging constitutes a major use case of FedXAI, where visual interpretability is often indispensable. Several studies address collaborative diagnosis tasks such as brain tumor classification~\cite{Mastoi2025Explainable}, skin cancer detection~\cite{Gupta2025FedMed-XAI,Naz2025Privacy}, eye disease diagnosis~\cite{KbFL2025mahadi}, and leukemia detection~\cite{Towards2025Pervez} using federated deep learning combined with visual explanation techniques. In these settings, Grad-CAM, saliency maps, and attention-based mechanisms are employed to highlight spatial regions that drive classification decisions, enabling clinicians to verify whether models attend to pathologically relevant areas. The integration of explainability is critical for building trust, particularly when models are trained across multiple institutions with heterogeneous data distributions.

Recent healthcare FedXAI frameworks further integrate advanced architectures and hybrid strategies to address data scarcity and generalization challenges. Examples include the use of GAN-based data augmentation and Vision Transformers for viral disease detection~\cite{Srinivasulu2025Overcoming,Mahir2026PriFL}, where explainability is provided through attention mechanisms and post-hoc attribution methods. Collectively, healthcare applications demonstrate that FedXAI can reconcile cross-institutional collaboration with privacy constraints while delivering explanations that are aligned with clinical reasoning.

A recent example is SHAP-FL proposed by Düsing and Cimiano~\cite{Dusing2025SHAPFL}, which addresses explanation inconsistency across healthcare institutions in federated sepsis onset prediction. 
The framework combines federated model training with a histogram-based background dataset synthesis mechanism that enables clients to generate more consistent SHAP explanations without sharing sensitive patient records. 
Using multicentric ICU data, the authors demonstrate that the synthesized federated background dataset improves both the fidelity and clinical plausibility of instance-level explanations across participating institutions. 
This work highlights an important emerging direction in healthcare FedXAI: harmonizing explanations across heterogeneous medical centers while preserving privacy.

However, most studies remain simulation-based, and systematic validation of explanation consistency and clinical utility across institutions remains an open challenge.

\subsection{Cybersecurity, Intrusion Detection, and IoT}
\label{subsec:fedxai-cybersecurity}

Cybersecurity and IDSs represent one of the most active and practically relevant domains of FedXAI. Network traffic data are highly sensitive, frequently encrypted, and distributed across heterogeneous administrative domains, making centralized learning impractical. At the same time, IDS outputs directly inform mitigation and response actions, rendering explainability a functional requirement rather than a purely diagnostic feature.

Several works integrate explainability into federated IDS pipelines to improve transparency and adaptability under non-IID data. Carillo \emph{et al.}~\cite{Explainable2025Carillo} combine FL with class-incremental learning for encrypted traffic classification, using attribution- and instance-based explanations to analyze how decision boundaries evolve across federated updates. Similarly, SHAP- and LIME-based explanations are widely used to justify intrusion alerts and audit feature relevance in edge-IoT environments~\cite{Kalakoti2025, Choir2016Sahoo,Fatema2025Federated}. In these systems, explainability supports both instance-level alert interpretation and model-level auditing.

A recent example is XAI-EdgeSFL proposed by Zhao \emph{et al.}~\cite{Zhao2026XAIEdgeSFL}, which integrates XAI with adaptive intrusion-resilient split FL for consumer healthcare IoT ecosystems. 
The framework combines lightweight edge-side Temporal Convolutional Networks with server-side classifiers, while employing SHAP, LIME, and Integrated Gradients to provide feature-level explanations for intrusion detection and health monitoring decisions. 
In addition to enhancing interpretability, the system introduces adaptive aggregation mechanisms to improve robustness against cyber threats under resource-constrained edge environments. 
This work illustrates an emerging direction in FedXAI that jointly addresses transparency, security, privacy preservation, and edge intelligence within healthcare-oriented IoT infrastructures.

FedXAI has also been applied to broader cyber-physical and industrial contexts. Yazdinejad \emph{et al.}~\cite{Explainable2025Yazdinejad} employ SHAP-based explanations to interpret threat detection decisions across cyber, physical, and social subsystems, while Patel \emph{et al.}~\cite{Demystifying2024Patel} use visual explanations to support fault diagnosis in semiconductor manufacturing. In addition, explainability has been used as an analytical tool to understand FL dynamics themselves, revealing how aggregation improves feature relevance consistency compared to isolated local training~\cite{Oki2024Evaluation}.

Overall, cybersecurity-oriented FedXAI emphasizes robustness, transparency, and operational trust. However, open challenges include explanation reliability under adversarial manipulation, the high computational cost of attribution methods, and the lack of standardized evaluation protocols for explanation robustness in federated IDS deployments.

\subsection{Energy, Smart Grid, and Transportation}
\label{subsec:fedxai-energy}

Energy systems, smart grids, and intelligent transportation networks form a key application domain of FedXAI, where privacy-preserving collaboration, operational efficiency, and interpretability are jointly required. Data in these domains are inherently distributed across geographically dispersed and independently operated entities, while AI-driven decisions often have direct economic and safety implications.

In smart grid applications, FL has been combined with explainable deep models for load forecasting, enabling collaborative training across grid nodes without exposing consumption data~\cite{Sarker2024Enhancing}. Explainability, typically implemented via SHAP or LIME, allows operators to identify the temporal and environmental factors driving demand fluctuations, supporting planning and anomaly detection. In renewable energy forecasting, explainability has been embedded more deeply into the learning process itself. Explainable Clustered FL (XCFL)~\cite{Explainable2025Ali} uses SHAP-based feature relevance not only for interpretation but also to guide client clustering and aggregation under strong geographical heterogeneity.

Transportation-oriented FedXAI further highlights the functional role of explainability in dynamic and resource-constrained environments. In vehicular networks and smart mobility systems, FL frameworks have been combined with explainability techniques to support privacy-preserving and adaptive decision-making for energy management, trajectory prediction, and intelligent transportation services.

Abdullah \emph{et al.}~\cite{Abdullah2025XFL} propose 
%an explainable federated learning (XFL) 
a FedXAI framework for autonomous electric vehicle energy management in smart cities, where FL enables decentralized optimization of vehicular energy consumption while XAI techniques enhance transparency and trustworthiness in AI-driven decisions. Their framework employs a hierarchical FL architecture using real-world vehicular telemetry data and integrates explainability mechanisms to interpret the influence of traffic density, energy consumption, and driving conditions on model predictions.

Similarly, weighted FL frameworks leverage SHAP- and LIME-based explanations to interpret energy consumption predictions and to adjust client contributions based on reliability and context~\cite{Weighted2025Saleem}.

Beyond energy optimization, federated and explainable approaches have also been explored for transportation-related IoT security and malicious traffic detection. Bilal \emph{et al.}~\cite{Bilal2025} integrate FL with SHAP and LIME for interpretable intrusion detection across distributed IoT and vehicular-edge environments, demonstrating how explainability can support transparent security monitoring while preserving data locality.

Across these applications, explainability supports both transparency and coordination, enabling federated systems to operate effectively under heterogeneous, non-IID, and rapidly changing transportation conditions. Nonetheless, scalability, communication overhead, and real-time explanation generation remain important open challenges in large-scale intelligent transportation deployments.

\subsection{Environment and Agriculture}
\label{subsec:fedxai-environment}

Environmental monitoring and smart agriculture are well suited to FedXAI due to the distributed nature of data sources, strong ownership constraints, and the high societal impact of predictive decisions. FL enables collaboration across farms, regions, or agencies without centralizing sensitive environmental data, while explainability is essential for trust and actionable insight.

In smart agriculture, FedXAI frameworks employ SHAP-based feature attribution to reveal how environmental factors such as soil moisture, temperature, rainfall, and nutrient levels influence predictions related to yield estimation and irrigation planning~\cite{Tahir2025Federated}. Explainability plays a decision-support role by translating model outputs into agronomically meaningful insights. In environmental risk modeling, FL combined with SHAP-based explanations has been applied to flood prediction in transboundary river basins~\cite{Mahir2025Advanced}. In this context, explainability highlights the relative contribution of upstream rainfall and hydrological dynamics, supporting policy-making and disaster preparedness.

Across these applications, explainability serves to align model behavior with physical and environmental understanding rather than to justify individual predictions. Open challenges include handling long-term temporal dependencies, validating explanations against domain-specific scientific models, and designing benchmarks that capture realistic cross-region dependencies and climate variability.

\subsection{Finance}
\label{subsec:fedxai-finance}

Financial fraud detection is a regulation-driven application domain where FedXAI is particularly compelling. Financial institutions are prohibited from sharing raw transaction data, yet are required to provide transparent and auditable explanations for automated decisions. FedXAI directly addresses this tension by enabling collaborative learning with built-in interpretability~\cite{popat2026federated}.

Several studies integrate SHAP-based explainability into federated fraud detection pipelines, allowing institutions to collaboratively train models while interpreting transaction-level risk factors~\cite{Aljunaid2025Secure}. In these systems, explainability supports regulatory compliance, post-hoc investigation, and internal risk assessment, rather than serving as a purely diagnostic tool. Feature-level explanations enable auditors and analysts to assess whether predictions are grounded in plausible financial behavior.

More conceptual analyses further emphasize that privacy-preserving learning alone is insufficient for trustworthy fraud detection, as accountability and interpretability are increasingly mandated by regulators~\cite{Transparency2024Awosika}. Across the financial domain, explainability functions as a trust-enabling mechanism that bridges technical fraud models and non-technical stakeholders. However, challenges remain in ensuring explanation consistency across institutions, mitigating privacy leakage through feature attributions, and validating FedXAI systems in real-world inter-bank deployments.

\section{Open Challenges and Future Directions (Roadmap)}

FedXAI has progressed from a conceptual vision to a rapidly expanding body of applied research across domains such as healthcare, cybersecurity, smart grids, and intelligent transportation systems. Despite these advances, the literature reveals that FedXAI is still in an early, pre-paradigmatic stage, characterized by fragmented methodologies, domain-specific solutions, and limited theoretical grounding. The combination of FL constraints with explainability requirements introduces challenges that are not encountered when these paradigms are studied in isolation, making the formulation of a coherent roadmap both necessary and timely.

A fundamental open challenge concerns the privacy of explanations themselves. While FL prevents direct data sharing, many FedXAI frameworks implicitly treat explanations as non-sensitive artifacts. In practice, however, feature attributions, saliency maps, and explanation statistics can encode sensitive information about local data distributions, rare events, or client-specific patterns. Several studies exchange or aggregate SHAP-based information to enable model-level interpretability or explanation-driven optimization, thereby introducing new leakage channels that are not covered by classical FL threat models. Although differential privacy and cryptographic mechanisms have been incorporated into some FedXAI pipelines, empirical results indicate a significant trade-off between explanation fidelity, predictive performance, and computational overhead~\cite{Explainable2025Yazdinejad,Tawfik2025FedMedSecure}. Foundational work explicitly highlights the absence of a formal notion of explanation privacy in FedXAI, underscoring the need for principled privacy models tailored to explanatory artifacts rather than model parameters alone~\cite{Corcuera2022Fed-XAI}.

Closely related is the largely unexplored issue of explanation integrity and manipulation. FedXAI systems are vulnerable not only to model poisoning but also to explanation poisoning, in which adversarial clients deliberately distort explanations while preserving high predictive accuracy. This threat is particularly severe in approaches where explanations actively influence learning dynamics, such as explanation-weighted aggregation, clustering, or resource management strategies~\cite{Explainable2025Ali,Patni2024Explainable}. While some security-oriented frameworks combine robust aggregation with explainability, explanation integrity is rarely treated as an explicit adversarial objective~\cite{Taheri2025Explainable,Chowdhury2026FLEX-IDS}. Addressing this gap requires new robustness mechanisms that jointly reason about prediction behavior and explanatory consistency.

Another major challenge arises from the interaction between non-IID data and explanation stability. non-IID data distributions are inherent to FL, yet FedXAI further demands that explanations remain meaningful, comparable, and stable across heterogeneous clients. Empirical evidence shows that feature attributions can vary substantially across clients and federated rounds, even when global performance converges. A small number of studies propose quantitative measures, such as attribution distance or explanation consistency, to analyze this phenomenon~\cite{Oki2024Evaluation,Bilal2025}. However, most works still rely on qualitative inspection, especially in complex multimodal or medical settings~\cite{Tahir2025Federated,Mastoi2025Explainable}. Future research should formalize explanation drift and stability as core evaluation dimensions and explore explanation-level personalization strategies that complement predictive personalization.

The problem becomes more pronounced in continual and streaming FedXAI scenarios. Real-world federated systems must adapt to evolving data distributions, emerging classes, and concept drift, yet the majority of FedXAI evaluations assume static datasets and limited training horizons. Preliminary results in federated class-incremental learning demonstrate that explainability can reveal catastrophic forgetting and bias shifts, but they also expose the fragility of explanations under continual updates~\cite{Explainable2025Carillo}. Conceptual studies identify streaming data and long-term accountability as central unresolved challenges in FedXAI~\cite{Corcuera2022Fed-XAI}. Consequently, future systems must move beyond explaining instantaneous predictions and instead support explanations of model evolution and temporal change.

Scalability and efficiency remain persistent bottlenecks. Post-hoc explainability techniques such as SHAP and LIME are computationally expensive and often dominate the runtime and communication cost of FedXAI pipelines, particularly in cross-device and edge-based deployments. Several works mitigate these costs through lightweight models, pruning, or selective explanation strategies, but these solutions are largely heuristic~\cite{Sarker2024Enhancing,Bilal2025}. The integration of cryptographic privacy mechanisms further exacerbates computational and communication overhead~\cite{Explainable2025Yazdinejad,Tawfik2025FedMedSecure}. This motivates research into explanation-efficient designs, including compressed or sparse explanations, on-demand interpretability, and architectural choices that reduce reliance on heavy post-hoc methods.

An alternative and promising research direction involves interpretable-by-design models in federated settings. Federated fuzzy rule-based systems and fuzzy regression trees demonstrate that strong interpretability, low communication cost, and competitive performance can be jointly achieved~\cite{Corcuera2022,Corcuera2025Increasing}. Tooling efforts such as OpenFL-XAI further show that such models can be operationalized in real federated workflows~\cite{OpenFL2023}. Nonetheless, these approaches face challenges related to rule complexity, aggregation conflicts, and scalability to high-dimensional or unstructured data. Hybrid architectures that combine deep representation learning with federated symbolic or rule-based explanation layers represent a compelling avenue for future research.

A notable gap across the literature is the limited emphasis on human-centered evaluation. Many FedXAI studies assess interpretability through visual plausibility or attribution rankings, yet few investigate whether these explanations improve human decision-making, calibration, or trust in real workflows. This limitation is particularly critical in high-stakes domains such as medical imaging and clinical decision support, where explanations are often cited as a prerequisite for deployment~\cite{Designing2022Raza,Mastoi2025Explainable,Manocha2025Federated}. Future work should incorporate task-based human evaluations, role-specific explanation design, and longitudinal studies of trust and reliance.

Finally, the lack of standardization in benchmarks, metrics, and evaluation protocols remains a major barrier to progress. There is no consensus on how to jointly assess accuracy, privacy, robustness, efficiency, and explanation quality in FedXAI systems. While some studies introduce ad hoc FedXAI-specific metrics such as attribution stability or calibration measures, these are rarely comparable across works~\cite{Choir2016Sahoo,Towards2025Pervez}. Foundational surveys explicitly identify this fragmentation as a key open challenge~\cite{Corcuera2022Fed-XAI}. Establishing standardized benchmarks, reference non-IID partitions, and unified reporting guidelines is essential for advancing FedXAI toward a mature and reproducible research field.

In summary, the future of FedXAI will be shaped less by incremental performance gains and more by the community's ability to secure explanations, ensure their integrity under adversarial and non-IID conditions, scale them efficiently, and validate their utility in human-centered decision-making contexts. Addressing these challenges will determine whether FedXAI evolves into a foundational paradigm for trustworthy federated intelligence or remains a collection of domain-specific case studies.

\section{Conclusion}

Federated Explainable Artificial Intelligence
(FedXAI)  has emerged as a critical research direction at the intersection of privacy-preserving learning and trustworthy artificial intelligence. By jointly addressing the decentralization constraints of Federated Learning (FL) and the transparency requirements of explainable AI, FedXAI seeks to enable collaborative intelligence in domains where both data confidentiality and human interpretability are non-negotiable. This survey has provided a comprehensive and structured analysis of this emerging paradigm, highlighting how explainability is no longer a peripheral add-on, but an increasingly integral component of federated system design.

A central contribution of this work is the introduction of a multi-axis taxonomy that systematically organizes the FedXAI literature along five orthogonal dimensions: the role of explainability within the FL lifecycle, the family of XAI techniques employed, the scope and target of explanations, the integration level and exchanged artifacts, and the underlying FL setting and heterogeneity assumptions. This taxonomy enables principled comparison across heterogeneous approaches and reveals fundamental design trade-offs between transparency, privacy, scalability, and model expressiveness. By moving beyond single-axis categorizations, the proposed framework captures the diversity of FedXAI methodologies ranging from post-hoc interpretability to explanation-aware optimization and interpretable-by-design federated models.

Through an extensive methodological review, this survey has identified three dominant paradigms in FedXAI: post-hoc explainability applied to federated models, explainability-aware FL where explanations actively shape coordination and aggregation, and interpretable-by-design approaches that embed transparency directly into the model structure. Each paradigm offers distinct advantages and limitations. Post-hoc methods provide flexibility and ease of deployment but remain passive and vulnerable to instability under non-IID data. Explainability-aware approaches demonstrate that explanation signals can improve robustness, personalization, and heterogeneity management, albeit at the cost of increased system complexity and new security risks. Interpretable-by-design models offer the strongest transparency guarantees and governance benefits, yet currently face scalability and applicability constraints in unstructured data domains.

The survey further highlights that evaluation remains a major bottleneck for the maturation of FedXAI. Existing studies rely on heterogeneous metrics, simulated federations, and qualitative assessments that limit reproducibility and cross-paper comparison. In particular, explanation quality, stability under non-IID data, privacy leakage from explanations, robustness against manipulation, and system-level overhead are rarely evaluated in a unified manner. The lack of standardized benchmarks and evaluation protocols underscores the need for community-wide efforts toward reproducible and application-relevant FedXAI assessment.

From an application perspective, the reviewed literature demonstrates that FedXAI is most mature in healthcare, cybersecurity, energy systems, and finance, where regulatory constraints and high-stakes decision-making necessitate both privacy and interpretability. However, most deployments remain experimental and simulation-based, indicating a gap between methodological innovation and real-world operational validation. Bridging this gap will require closer integration of domain knowledge, human-centered evaluation, and system-level design considerations.

In conclusion, FedXAI represents a paradigm shift in the design of distributed intelligent systems, reframing explainability as a first-class constraint alongside privacy and performance. While significant progress has been made, the field remains fragmented and faces open challenges related to explanation security, non-IID stability, scalability, and evaluation standardization. Addressing these challenges will be essential for transforming FedXAI from a collection of promising techniques into a foundational framework for trustworthy, transparent, and privacy-preserving federated intelligence.

\section*{Acknowledgements}
This work has been partly funded by the Italian Ministry of University and Research (MUR) within the framework of the PNRR - M4C2 - Investimento 1.3, Partenariato Esteso PE00000013 - ``FAIR - Future Artificial Intelligence Research'' - Spoke 1 ``Human-centered AI'' under the NextGeneration EU programme, and the framework of the FoReLab and CrossLab projects (Departments of Excellence).

\section*{Data Availability}
No new data were generated or analyzed in this study. All data discussed in this article are derived from previously published studies.

%\label{app1}



\bibliographystyle{elsarticle-num}
\bibliography{references}

\end{document}